\documentclass[12pt, draftclsnofoot, onecolumn]{IEEEtran}
\usepackage{subfigure}
\usepackage{caption}
\usepackage{graphicx}
\usepackage{amsthm}
\usepackage{epsfig}
\usepackage{latexsym}
\usepackage{amsfonts}
\usepackage{here}
\usepackage{rawfonts}
\usepackage[latin1]{inputenc}
\usepackage[T1]{fontenc}
\usepackage{calc}
\usepackage{capitalgreekitalic}
\usepackage{url}
\usepackage{enumerate}
\usepackage{color}
\usepackage[tbtags]{amsmath}
\usepackage{amssymb}
\usepackage{upref}
\usepackage{epic,eepic}
\usepackage{times}
\usepackage{dsfont}
\usepackage{comment}
\usepackage{cite}
\usepackage{bbm} 
\usepackage{bm}

\newtheorem{lemma}{\bf Lemma}

\usepackage{dsfont}

\newcounter{step}
\newlength{\totlinewidth}
\newenvironment{algorithm}{%
  \begin{list}{}%
    {\usecounter{step}%
      \settowidth{\labelwidth}{\textbf{Step 2:}}%
      \raggedright}}%
  {\end{list}%
  \rule{\linewidth}{1pt}}
\newcounter{substep}

  {\end{list}}

\IEEEoverridecommandlockouts

\usepackage{algorithm}
\usepackage{algorithmic}

\usepackage{subfigure}

\usepackage{enumitem}

\begin{document}
\captionsetup[figure]{labelformat=simple, labelsep=period}
\title{\huge Optimization of Image Transmission in a Cooperative Semantic Communication Networks \vspace*{0.5em}}
\author{{Wenjing Zhang, \emph{Student Member, IEEE}}, Yining Wang, \emph{Student Member, IEEE}, Mingzhe Chen, \emph{Member, IEEE}, Tao Luo, \emph{Senior Member, IEEE}, and Dusit Niyato, \emph{Fellow, IEEE} \vspace*{-2em}\\ 
\thanks{W. Zhang, Y. Wang, and T. Luo are with the Beijing Laboratory of Advanced Information Network, Beijing University of Posts and Telecommunications, Beijing, 100876, China (e-mail \protect\url{zhangwenjing@bupt.edu.cn}; \protect\url{wyy0206@bupt.edu.cn}; \protect\url{tluo@bupt.edu.cn}).}
\thanks{M. Chen is with the Department of Electrical and Computer Engineering and Institute for Data Science and Computing, University of Miami, Coral Gables, FL, 33146 USA (Email: mingzhe.chen@miami.edu).}
\thanks{D. Niyato is with the School of Computer Science and Engineering (SCSE), NTU, Singapore (e-mail: \protect\url{dniyato@ntu.edu.sg}).}
\thanks{A preliminary version of this work \cite{ZWJ} is accepted by the Proceedings of the 2022 IEEE International Global Communications Conference (GLOBECOM)}
 }
\maketitle
%
\begin{abstract}In this paper, a semantic communication framework for image data transmission is developed. In the investigated framework, a set of servers cooperatively transmit image data to a set of users utilizing semantic communication techniques, which enable servers to transmit only the semantic information that accurately captures the meaning of images. To evaluate the performance of studied semantic communication system, a multimodal metric called image-to-graph semantic similarity (ISS) is proposed to measure the correlation between the extracted semantic information and the original image. To meet the ISS requirement of each user, each server must jointly determine the semantic information to be transmitted and the resource blocks (RBs) used for semantic information transmission. Due to the co-channel interference among users associated with different servers, each server must cooperate with other servers to find a globally optimal semantic oriented RB allocation. We formulate this problem as an optimization problem whose goal is to minimize the sum of the average transmission latency of each server while reaching the ISS requirement. To solve this problem, we propose a value decomposition based entropy-maximized multi-agent reinforcement learning (RL) algorithm. The proposed algorithm enables each server to coordinate with other servers in training stage and execute RB allocation in a distributed manner to approach to a globally optimal performance with less training iterations. Compared to traditional multi-agent RL algorithms, the proposed RL framework improves the exploration of valuable action of servers and the probability of finding a globally optimal RB allocation policy based on local observation of wireless and semantic communication environments. Simulation results show that the proposed algorithm can reduce the transmission delay by up to 16.1\% and improve the convergence speed by up to 100\% compared to the traditional multi-agent RL algorithms.          
\end{abstract}


\section{Introduction}
Current communication technologies are trying to approach the Shannon physical capacity limit \cite{CMZ,MXD,ZGX}. 
The integration of communication and artificial intelligence (AI) technology promotes the development of communication to a higher level, i.e., from the technical level to the semantic level\cite{Walid,KM,KJW}. A paradigm called \emph{semantic communication}, shifts from rate-centric towards content-aware communication technologies has been proposed \cite{Qin,LXW,LZY,ZH}, to effectively transmit a fast-growing amount of data (i.e., image, video, and immersive data) over wireless networks \cite{MingzheChen,Tandon,ZM}. Semantic communications enable devices to communicate with each other using the desired meaning of the original data so as to improve communication efficiency \cite{Shi,HTX,JPW}. However, current semantic communication techniques are mostly studied for text and image data transmission. Compared to textual data where semantic information is explicitly represented by words, semantic information in an image is implicit. Therefore, developing a semantic communication framework for image transmission faces several challenges including: 1) human-oriented semantic information representation, 2) metric design for image semantic information, and 3) dynamic semantic information extraction based on users' service requirements.

\subsection{Related Works}

Recently, semantic communications over wireless networks have been studied in \cite{Bao, Basu, zhw, YY, Wangyining, Viass}. In \cite{Bao}, the authors investigated a logistic probability based semantic information measurement. In \cite{Basu}, the authors defined the semantic channel capacity of a semantic communication system as mutual information between semantic information and original data. However, both metrics designed in \cite{Bao} and \cite{Basu} measure only the received semantic information with logistic true without considering the completeness of the meaning that is expressed by the semantic information. The authors in \cite{zhw} and \cite{YY} investigated a deep learning based semantic communication system that compresses original data into vectors and considers the compressed vectors as semantic information. However, these vectors do not have any practical meanings and are incomprehensible for human receivers. The authors in \cite{Wangyining} introduced a text semantic communication framework that seeks to maximize the semantic similarity between original data and semantic information. The authors in \cite{Viass} used the accuracy of the receive semantic information to measure the performance of the proposed semantic communication system. However, the metrics defined in \cite{Wangyining} and \cite{Viass} are based on the consistency of textual data in a word level, which cannot be used for image data. 

The works in \cite{Xie, Compress, Dan, LX} studied the use of semantic communication techniques for image transmission. In particular, the works in \cite{Xie} and \cite{Compress} designed an image semantic communication system aiming to improve image compression ratio. The authors in \cite{Dan} introduced an image semantic coding model and defined a rate-perception-distortion metric to evaluate the performance of the proposed model. The authors in \cite{LX} investigated a task-driven semantic coding framework of image. However, these works in \cite{Xie, Compress, Dan, LX} modeled the semantic information of an image as uninterpretable feature vectors that cannot be directly utilized and understood by human receivers. Hence, the receivers in these works \cite{Xie,Compress,Dan, LX} need to reconstruct original images, which is inefficient and complicated since the receivers need to use neural networks to interpret received data into explainable and meaningful information. 

Currently, a number of existing works studied the use of RL for semantic communication performance optimization. In particular, the authors in \cite{Wangyining} utilized an attention-based RL algorithm to analyze the relationship between the original data and its semantic information. The authors in \cite{Viass} investigated a self-critic policy gradient enabled semantic communication system. The works in \cite{Dan} designed an RL based adaptive semantic coding model. The works in \cite{LX} utilized RL to determine the quantization parameters of semantic coding in different tasks. However, these works do not consider the cooperation among different agents and hence each agent's performance will be affected by the actions of other agents thus reducing network performance achieved by RL. The authors in \cite{icassp} used a value decomposition based deep Q-learning network (DQN) to reduce transmission delay and energy consumption in a semantic communication based network. However, DQN related RL requires a large amount of users' historical experience due to its weak exploration ability to find a globally optimal solution. 

\subsection{Contributions}
The main goal of this work is to design a novel image semantic communication framework that enables a set of servers to cooperatively transmit images to users using semantic communication techniques. The key contributions include:

\begin{itemize}
\item We consider a semantic communication system in which a set of servers collaboratively transmit image data to a set of users using semantic communication techniques. The semantic information extracted from an image is modeled by a scene graph (SG) that captures the objects and their relationships in the original image. 
\item To evaluate the semantic similarity between the semantic information and its original image, we introduce a comprehensive multimodal image-to-graph semantic similarity (ISS) metric. Compared to conventional metrics such as structural similarity (SSIM) that measures the differences in a set of pixels, ISS can capture the correlation of the meaning between the original image and its semantic information.
\item To meet the target ISS requirement of each user, each server must jointly determine the partial semantic information to be transmitted and resource blocks (RBs) used for semantic information transmission. We formulate this problem as an optimization problem whose goal is to minimize the sum of the average transmission latency of all users while meeting the ISS requirement. 
\item To solve the optimization problem, we propose a novel value decomposition based entropy-maximized multi-agent deep reinforcement learning (VD-ERL) algorithm. Compared to traditional multi-agent RL \cite{icassp} and \cite{IQL}, the proposed algorithm enables servers to achieve globally optimal performance with less training iterations. Meanwhile, the proposed algorithm can improve the action exploration and the probability of finding a near optimal cooperative RB allocation policy. 
\end{itemize}
Simulation results show that, compared to traditional multi-agent RL algorithms, the proposed VD-ERL algorithm can reduce the transmission delay by up to 16.1\% while reducing 50\% iterations to converge. \emph{To the best of our knowledge, this is the first work that introduces an image semantic communication framework which jointly optimizes the RB allocation of multi-server to minimize the sum of the average transmission latency of all users while satisfying the ISS requirement.}

The rest of this paper is organized as follows. The proposed image semantic communication system model and the problem formulation are described in Section \uppercase\expandafter{\romannumeral2}. Section \uppercase\expandafter{\romannumeral3} introduces the proposed VD based entropy-maximized multi-agent RL for cooperative semantic-oriented RB allocations. In Section \uppercase\expandafter{\romannumeral4}, numerical results are presented and discussed. Finally, conclusion are drawn in Section \uppercase\expandafter{\romannumeral5}.

\begin{figure}
\centering
\includegraphics[width=13cm]{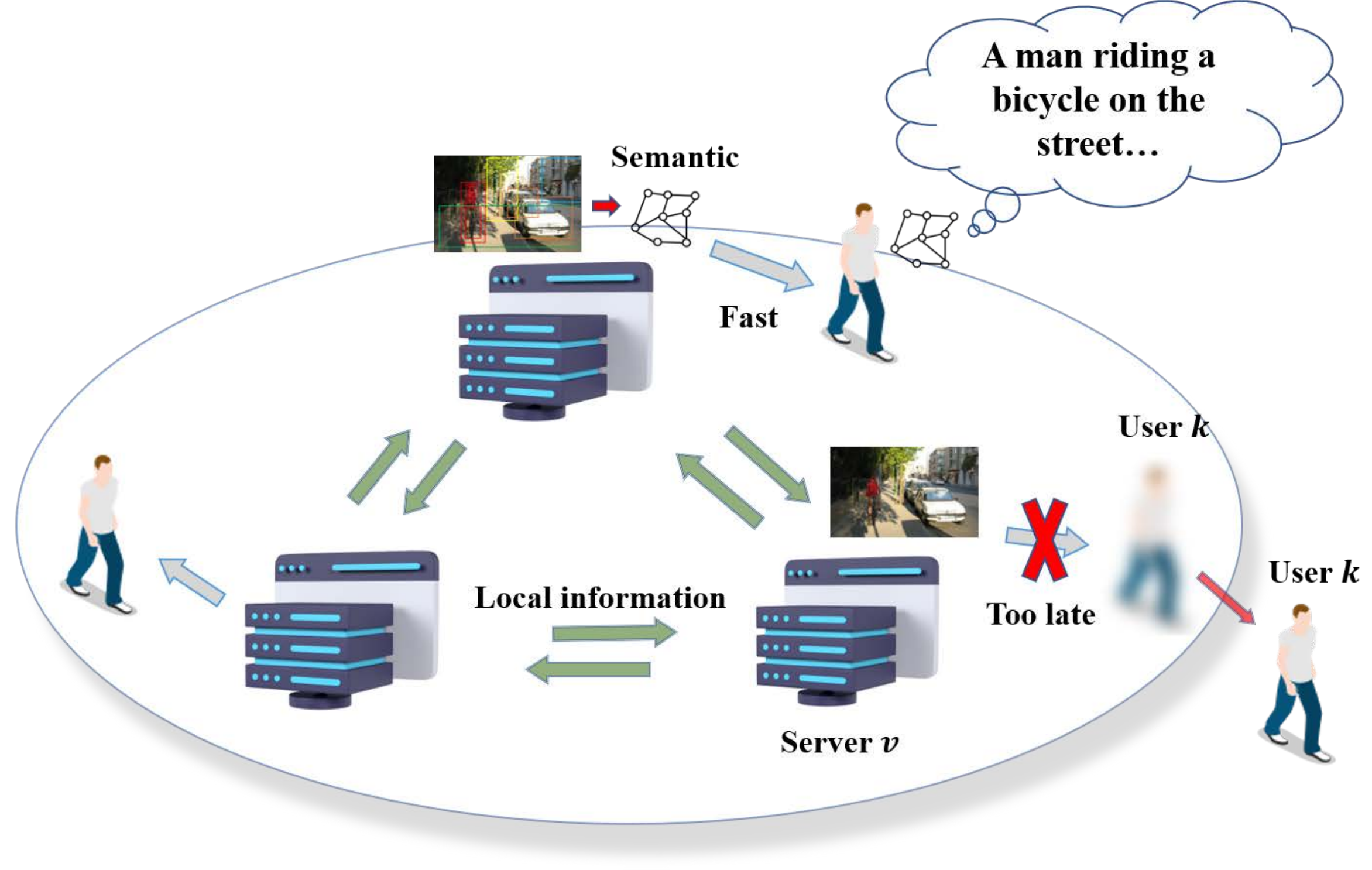}
\caption{The cooperative multi-server image semantic communication wireless network.}\label{fig1}
\end{figure}
\begin{figure}[t]
\centering
\includegraphics[width=12cm]{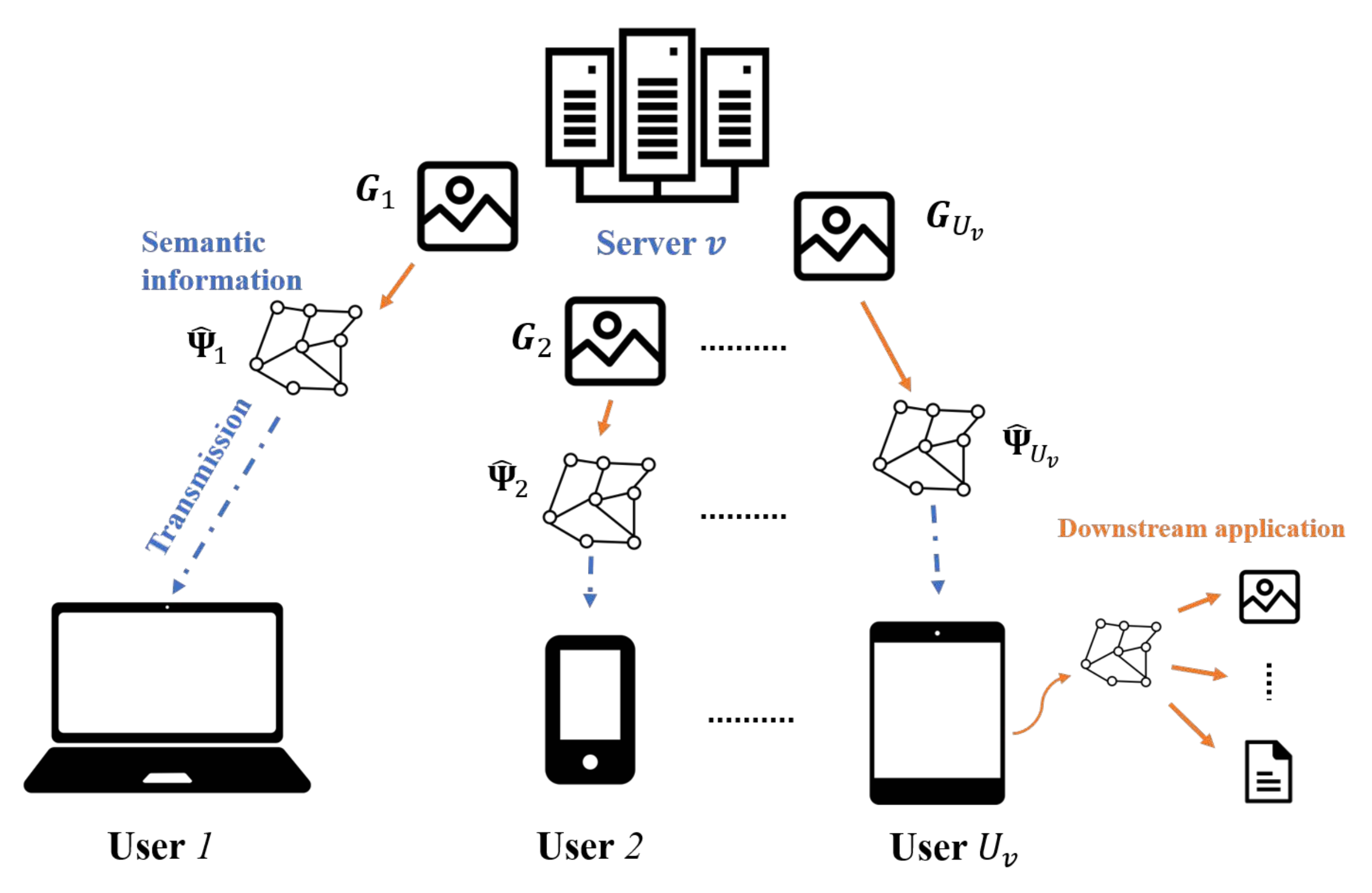}
\caption{The image semantic transmission framework of each server.}
\label{fig2}
\end{figure}

\begin{table}\centering\footnotesize
\setlength{\belowcaptionskip}{0pt}
\setlength{\abovedisplayskip}{-15pt}
\setlength{\tabcolsep}{1.5mm}{
\newcommand{\tabincell}[2]{\begin{tabular}{@{}#1@{}}#2.0\end{tabular}}
\renewcommand\arraystretch{1.5}
\caption[table]{{List of notations}}
\label{table:notations}
\centering
\begin{tabular}{|c|c|c|c|}
\hline
\!\textbf{Notation}\! \!\!& \!\!\!\!\textbf{Description} &\!\textbf{Notation}\! \!\!& \!\!\!\!\textbf{Description} \\
\hline
$V$ & Number of servers & $U$ & Number of users     \\
\hline
$Q$ & Number of downlink orthogonal RBs & $W$ & Bandwidth of each RB \\
\hline
$P$ & Transmit power of the server & $N_0$ & Noise power spectral density \\
\hline
$I_{vk}^q$ & Interference of RB $q$ & $h_{vk}^q$ & Channel gain of RB $q$  \\
\hline
$\bm{a}_{k}$ & RB allocation vector of user $k$ & $c_{k}\left(\bm{a}_{k}\right)$ & Downlink channel capacity of user $k$\\
\hline
 $G_{k}$ & Original image needed to transmit to user $k$ & $e_{vk,i}$ & Object $i$ in $G_{k}$  \\
\hline
$l_{vk,ij}$ & Relationship between objects $e_{vk,i}$ and $e_{vk,j}$ & $\bm{\Psi}_{vk}$ & Semantic information of image $G_{k}$  \\
\hline
$\bm{\psi}_{vk}^n$ & Semantic triple $n$ in $\bm{\Psi}_{vk}$ & $Z\left(\bm{\Psi}_{vk}\right)$ & Number of letters in $\bm{\Psi}_{vk}$ \\
\hline
$\bm{\hat{\Psi}}_{vk}$ & Transmitted semantic information & $\hat{N}_{vk}$ & Number of semantic triples in $\bm{\hat{\Psi}}_{vk}$  \\
\hline
$\epsilon$ & Semantic reliability threshold & $\xi$ & Minimum acceptable semantic similarity \\
\hline
$E\left(\bm{\hat{\Psi}}_{vk},\bm{a}_{vk}\right)$ & Image-to-graph semantic similarity & $T\left(\bm{\hat{\Psi}}_{vk},\bm{a}_{vk}\right)$ & Transmission latency of user $k$  \\
\hline
$C\left(G_{k}\right)$ & vectorized image $G_{k}$ & $\bm{O}_{vk}$ & vectorized partial semantic information $\bm{\hat{\Psi}}_{vk}$ 
\\
\hline
$\rho$ & penalty of failed association & &\\
\hline
\end{tabular}}
\end{table}

\section{System Model and Problem Formulation}\label{se:system}
Consider a cellular network in which a set $\cal{V}$ of $V$ servers cooperatively transmit image data to a set $\cal{U}$ of $\emph{U}$ users using semantic communication techniques, as shown in Fig.~\ref{fig1}. Let ${\cal L}_v$ represent a set of the users that are located in the service area of server $v$. Here, the service areas of different servers may overlap. The procedure of the considered semantic communication of each server consists of two phases (as shown in Fig.~\ref{fig2}): a) semantic information extraction and b) semantic information transmission. Next, we introduce the process of the semantic information extraction. Then, we present a multimodal metric for the proposed image semantic communication framework which can evaluate the semantic similarity between the original image and its extracted semantic information. Table \ref{table:notations} summarizes all parameters used in our work.

\subsection{Semantic Information Extraction}
\begin{figure}[t]
\centering  
\subfigure[Input image.]{
\centering
\includegraphics[width=10cm]{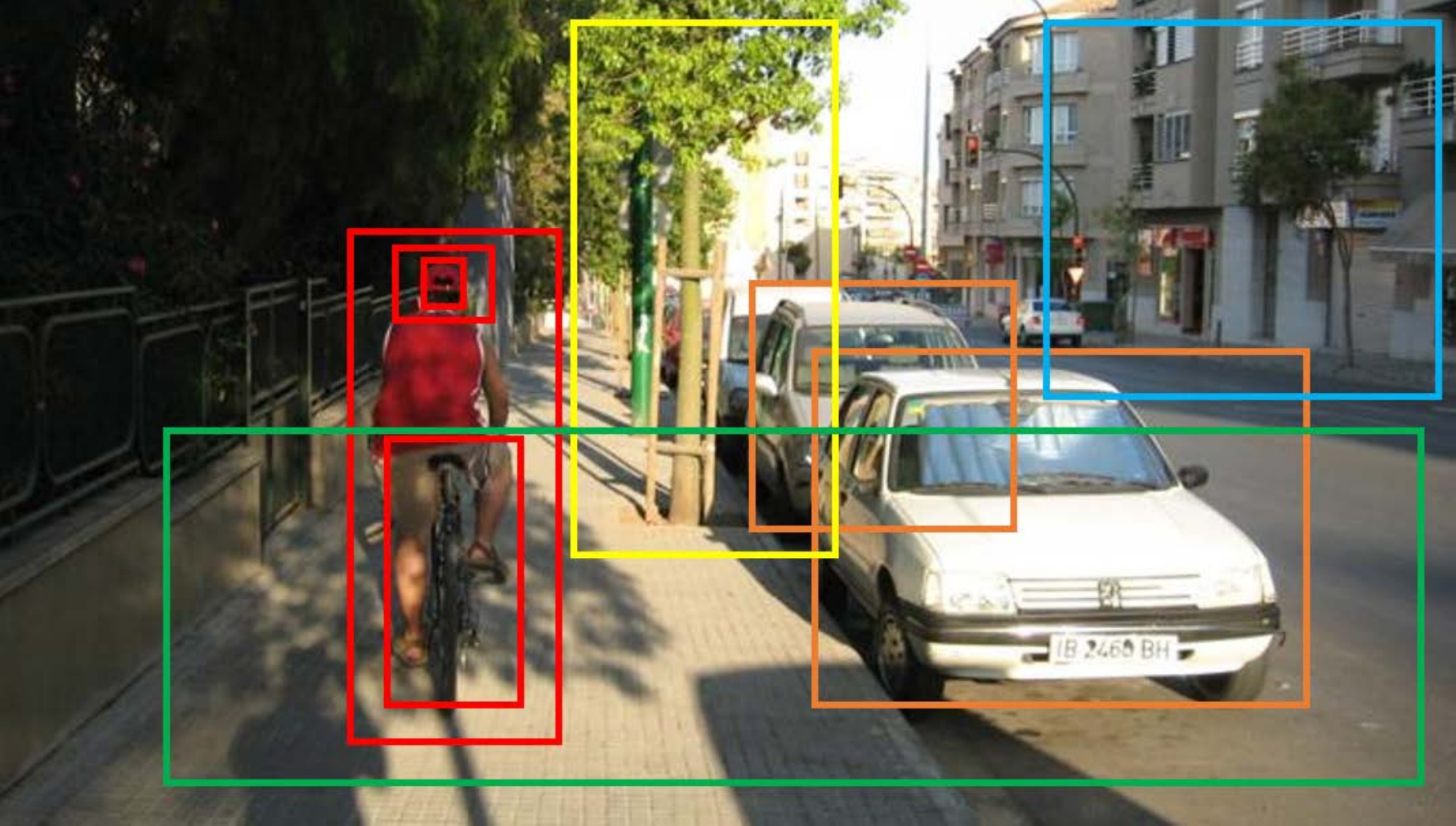}
}
\subfigure[The extracted semantic information.]{
\centering
\includegraphics[width=10cm]{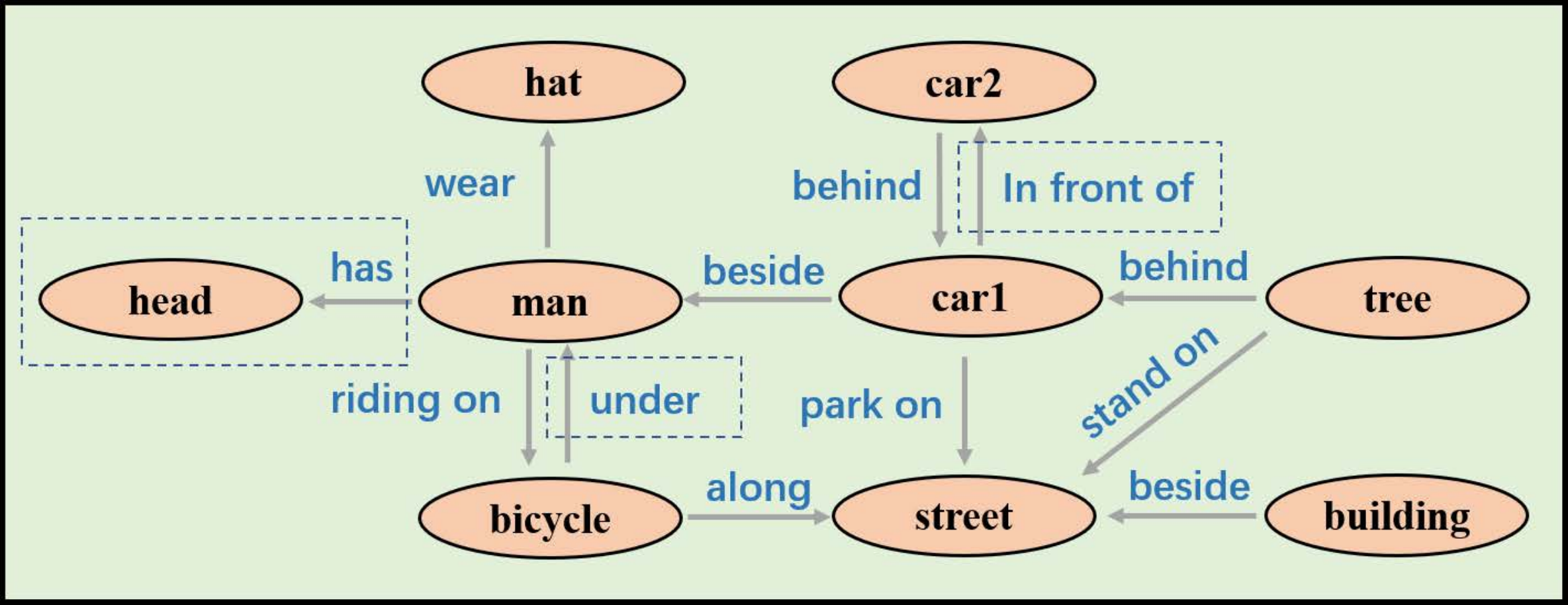}
}
\subfigure[The transmitted semantic information.]{
\centering
\includegraphics[width=10cm]{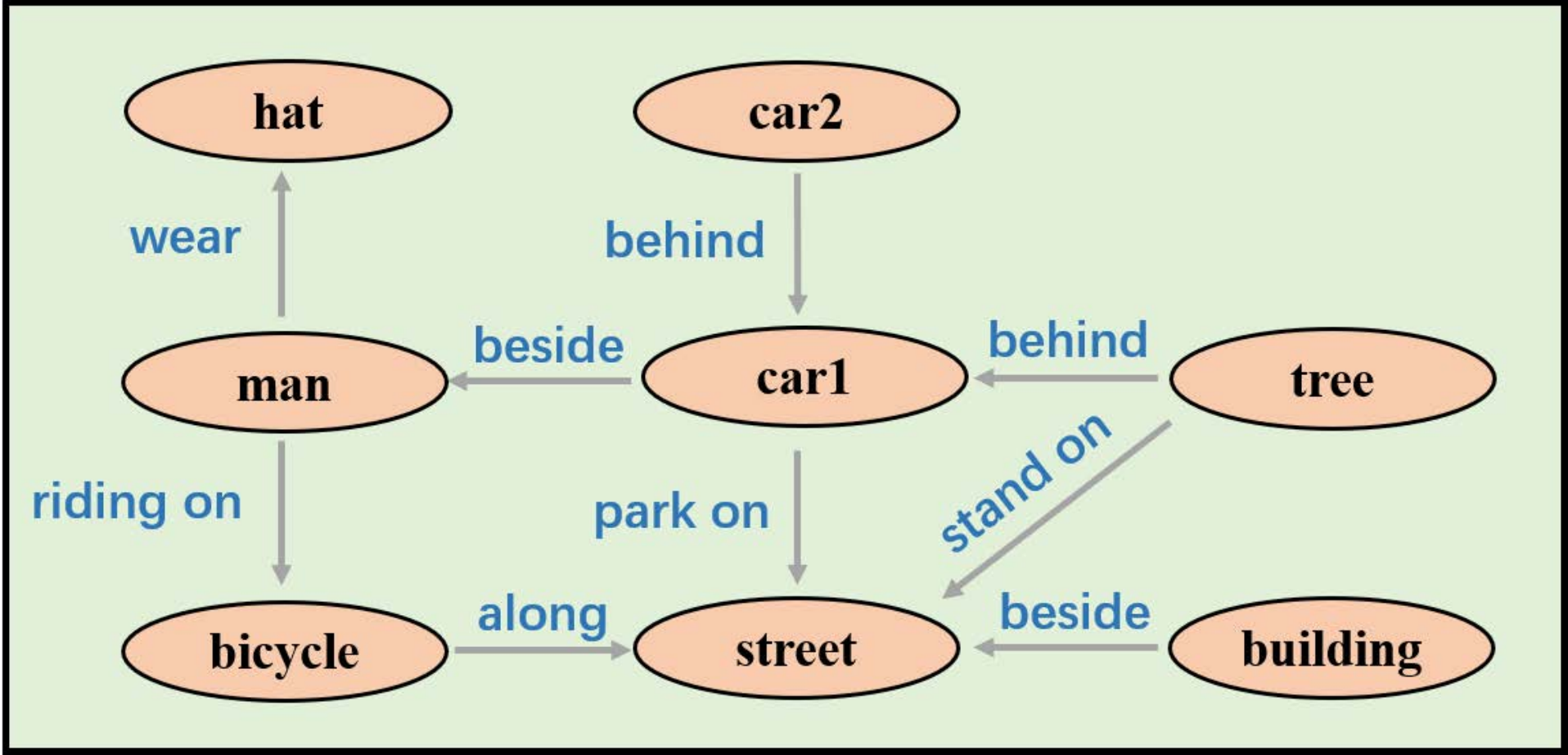}
}
\caption{An example of semantic information extraction.}
\label{fig3}
\end{figure}

In our model, we assume that the semantic information of an image consists of the objects and their relationships in the image. Hence, the semantic information of each image is modeled by a scene graph defined by a set of nodes and edges, where a node represents an object (e.g., \emph{a man}) and an edge represents the relationship between two objects, as shown in Fig.~\ref{fig3}. The semantic triple is a basic component of semantic information, which consists of two objects and the relationship between them. For example, a semantic triple in Fig.~\ref{fig3} is ([``\emph{man}''], [``\emph{riding on}''], [``\emph{bicycle}'']), where [``\emph{man}''] and [``\emph{bicycle}''] are objects and [``\emph{riding on}''] is their relationship. An image that a server needs to transmit can be described by multiple semantic triples.

The semantic information extraction process has two steps which are object identification and relationship capture. First, a server detects the region of the objects and identify their categories. Then, according to the geometry and logical correlation between the objects and their categories, the relationship between two objects can be captured by using a deep neural network model \cite{Vctree,Vsg,TDE}. The semantic information of an image $G_{k}$ that is extracted by server $v$ and transmitted to user $k$ can be expressed as 
\begin{equation}
\bm{\Psi}_{vk}  =\left\{\bm{\psi}^1_{vk},\bm{\psi}^2_{vk},\ldots,\bm{\psi}^n_{vk},\ldots,\bm{\psi}^{N_{vk}}_{vk}\right\},
\end{equation}
where $\bm{\psi}^n_{vk} = \left(e^n_{vk,i}, l^n_{vk,ij},e^n_{vk,j}\right)$ is a semantic triple and $N_{vk}$ is the number of semantic triples in image $G_{k}$, $e^n_{vk,i}$ is the category of object $i$ in image $G_{k}$, $l^n_{vk,ij}$ is the relationship between objects $e^n_{vk,i}$ and $e^n_{vk,j}$. Here, $l^n_{vk,ij}$ is directional and hence, $l^n_{vk,ij} \neq l^n_{vk,ji}$. To measure the size of the semantic information, we define $Z\left(\bm{x}\right)$ as the number of letters in word sequence $\bm{x}$. Therefore, the total number of letters in each image semantic information $\bm{\Psi}_{vk}$ is
\begin{equation} \label{$Z$}
Z\left(\boldsymbol{\Psi}_{vk}\right)  =\sum_{n=1}^{N_{vk}}\left(Z\left(e^n_{vk,i}\right)+Z\left(l^n_{vk,ij}\right)+Z\left(e^n_{vk,j}\right)\right).
\end{equation}
For example, in Fig.~\ref{fig3}, the number of letters in semantic triple $\bm{\psi}_{vk}^n$ = ([``\emph{man}''], [``\emph{riding on}''], [``\emph{bicycle}'']) is $Z\left(\bm{\psi}_{vk}^n\right)=Z\left(e^n_{vk,i}\right)+Z\left(l^n_{vk,ij}\right)+Z\left(e^n_{vk,j}\right)=3+8+7=18$. 

Note that some semantic triples in $\bm{\Psi}_{vk}$ may not contain useful information. For example, in Fig.~\ref{fig3}, we do not want to transmit the meaningless semantic triples such as ([``\emph{man}''], [``\emph{has}''], [``\emph{head}'']) and redundant semantic triples such as ([``\emph{bicycle}''], [``\emph{under}''], [``\emph{man}'']). In order to improve the efficiency of the considered semantic communication model, as shown in Fig. \ref{fig3}c), each server $v$ must transmit the semantic triples that contain the most significant meaning of an image. The partial semantic information that server $v$ transmits to user $k$ can be given as

\begin{equation}
\boldsymbol{\hat{\Psi}}_{vk}  =\left\{\bm{\hat{\psi}}^1_{vk},\bm{\hat{\psi}}^2_{vk},\ldots,\bm{\hat{\psi}}^n_{vk},\ldots,\bm{\hat{\psi}}^{\hat{N}_{vk}}_{vk}\right\}\subset \boldsymbol{\Psi}_{vk} ,
\end{equation}
where $\hat{N}_{vk}$ is the number of selected semantic triples in $\bm{\hat{\Psi}}_{vk}$.

\subsection{Transmission Model}
We assume that an orthogonal frequency division multiple access (OFDMA) technique is adopted. A set $\cal{Q}$ of Q downlink orthogonal RBs are allocated to serve users. The servers can reuse all these RBs and thus each server can allocate Q RBs to its associated users. The downlink rate of a server transmitting partial semantic information $\bm{\hat{\Psi}}_{vk}$ to user $k$ is given as 
\begin{equation}
c_{k}\left(\bm{a}_k\right)=\sum_{v=1}^V\sum_{q=1}^Q{a_{vk}^qWlog_2\left(1+\frac{Ph_{vk}^q}{I_{vk}^q+WN_0}\right)},
\end{equation}
where $P$ is the transmit power of server $v$, $W$ is the bandwidth of RB $q$ which is assumed to be equal for all RBs, $h_{vk}^q=\gamma_{vk}^q d_{vk}^{-2}$ is the channel gain between server $v$ and user $k$ with $\gamma_{vk}^q$ being the Rayleigh fading parameter and $d_{vk}$ being the distance between server $v$ and user $k$, $I_{vk}^q=\sum_{s\in {\cal V}_q,s\neq v}Ph_{s_k}^q$ represents the interference caused by other servers with ${\cal V}_q$ being the set of servers that use RB $q$, $N_0$ is the noise power spectral density, and $\bm{a}_k=[\bm{a}_{1k},\ldots,\bm{a}_{vk},\ldots,\bm{a}_{Vk}]$ with $\bm{a}_{vk}=\left[a_{vk}^1,\ldots,a_{vk}^Q\right]$ is an RB allocation vector for user $k$ of server $v$ with $a_{vk}^q\in \left\{0,1\right\}$ being the user-server connection index. In particular, $a_{vk}^q=1$ implies that server $v$ transmits semantic information to user $k$ using RB $q$, and $a_{vk}^q=0$, otherwise. Here,  each user can only be served by one server with one RB, and each RB of a server can only be allocated to one user. Then, we have $\sum_{v=1}^V\sum_{q=1}^Qa_{vk}^q\leqslant1,\forall k \in {\cal U}$ and $\sum_{k\in {\cal U}}a_{vk}^q\leqslant1 ,\forall v \in {\cal V},\forall q \in {\cal Q}$. According to (2), (3), and (4), the transmission latency of server $v$ transmitting selected partial semantic information $\bm{\hat{\Psi}}_{vk}$ to user $k$ can be given as
\begin{equation}
T\left(\bm{\hat{\Psi}}_{vk},\bm{a}_{vk}\right)= \frac{Z\left(\bm{\hat{\Psi}}_{vk}\right)}{c_{vk}\left(\bm{a}_{vk}\right)},
\end{equation}
where $c_{vk}\left(\bm{a}_{vk}\right)=\sum_{q=1}^Q{a_{vk}^qWlog_2\left(1+\frac{Ph_{vk}^q}{I_{vk}^q+WN_0}\right)}$ is the transmitting rate. Here, we note that only the transmission latency of associated user are considered and calculated. From (5), we see that the transmission latency of semantic information depends on user association, RB allocation, and the data size of the transmitted partial semantic information. Hence, for a certain user, if its associated server changes, its received semantic information extracted from the same image will be different. Moreover, changes of the wireless communication environment such as dynamic channel will affect its received semantic information.

\subsection{Image Semantic Similarity Model}
To evaluate the performance of image semantic communication, we propose a metric called image-to-graph semantic similarity (ISS). Different from conventional metrics, such as structural similarity (SSIM) \cite{SSIM}, that measure the differences in a set of pixels, the proposed metric can capture the correlation of the meaning between the extracted semantic information and its original image. We first use a deep neural network (DNN) based encoder to vectorize original image $G_{k}$ and the semantic information $\bm{\hat{\Psi}}_{vk}$ received by user $k$. The vectorized image data is $\bm{C}\left(G_{k}\right)$ and the vectorized semantic information is $\bm{O}_{vk}=\left\{\bm{C}\left(\bm{\hat{\psi}}_{vk}^1\right),\ldots,\bm{C}\left(\bm{\hat{\psi}}_{vk}^n\right),\ldots,\bm{C}\left(\bm{\hat{\psi}}_{vk}^{\hat{N}_{vk}}\right)\right\}$, where $\bm{C}\left(\cdot\right)$
is the vectorization function that constructs the relationship between the input semantic information and image by matching the text-image pairs with similar meaning.

The proposed ISS metric is defined as the included angle cosine between an image vector and its normalized semantic triple vectors, which is calculated by the projection of image vector on semantic information vector set. To build the basis of the semantic information vector set, the Gram-Schmidt algorithm is used to orthogonalize the semantic information vectors, which is given by $\overline{\bm{O}_{vk}}=\{\overline{\bm{C}(\bm{\hat{\psi}}_{vk}^1)},\ldots,\overline{\bm{C}(\bm{\hat{\psi}}_{vk}^n)},\ldots,\overline{\bm{C}(\bm{\hat{\psi}}_{vk}^{\hat{N}_{vk}})}\}$. Then, the ISS of semantic information $\bm{\hat{\Psi}}_{vk}$ that transmitted from server $v$ to user $k$ is defined as:
\begin{equation}
E\left(\bm{\hat{\Psi}}_{vk},\bm{a}_{vk}\right)=\left(\sum_{q=1}^{Q}a_{vk}^q\right)\frac{\Vert \sum\limits_{n=1}^{\hat{N}_{vk}}{{\lvert \overline{\bm{C}\left(\bm{\hat{\psi}}_{vk}^n\right)}\cdot \bm{C}\left(G_{k}\right)^T}\rvert}\overline{\bm{C}\left(\bm{\hat{\psi}}_{vk}^n\right)} \Vert }{ \Vert \bm{C}\left(G_{k}\right) \Vert}.
\end{equation}
From (6), we see that the value of the ISS increases as the number of transmitted semantic triples increases, which is consistent with the objective human cognition \cite{hvs}.

In the proposed framework, each server $v$ transmits only partial semantic information and, hence, the received semantic information includes a part of meaning of the image. We define the minimum acceptable ISS of each user as $\xi$. Then, the probability of the received partial semantic information satisfying $E\left(\boldsymbol{\hat{\Psi}}_{vk},\bm{a}_{vk}\right) \geqslant \xi$ is defined as the semantic reliability, which is given by   
\begin{equation}
{P\left(E\left(\boldsymbol{\hat{\Psi}}_{vk},\bm{a}_{vk}\right) \geqslant \xi\right) \geqslant \epsilon},
\end{equation}
where $\epsilon$ is the semantic reliability threshold that is used to adjust the probability of reliable semantic transmission. For example, $\xi=0.6$ and $\epsilon=0.9$ represents that at least 90\% semantic information transmission must satisfy $E\left(\boldsymbol{\hat{\Psi}}_{vk},\bm{a}_{vk}\right)\geqslant 0.6$.

\subsection{Problem Formulation}
Given the defined system model, our objective is to minimize the average transmission latency of all users while satisfying the semantic reliability requirement. This minimization problem includes optimizing the user association, RB allocation, and determining the part of semantic information to transmit. The average transmission latency minimization problem is formulated as follows: 
\begin{subequations}\label{eq:optimal_problem}
\begin{align}\tag{8}
&{\;\;\;\;\mathop {\min }\limits_{\bm{\hat{\Psi}}_{vk},\bm{a}_{vk}} \!\frac{\sum_{v\in {\cal V}}\sum_{k\in {\cal U}_v} \; T\left(\bm{\hat{\Psi}}_{vk},\bm{a}_{vk}\right)}{\sum_{v\in {\cal V}}\vert {\cal U}_v\vert}}\\
&{\;\;\;\;\;\;{\rm{s}}.{\rm{t}}.\;\;\;\;\;{a_{vk}^q\in\{0,1\}},{\forall k \in {\cal U}_v}, \forall v \in {\cal V}, \forall q \in {\cal Q}},\\
&{\;\;\;\;\;\;\;\;\;\;\;\;\;\;\;\;\sum_{v\in{\cal V}}\sum_{q\in{\cal Q}}a_{vk}^q\! \leqslant1 , \forall k \in {\cal U}_v},\\
&{\;\;\;\;\;\;\;\;\;\;\;\;\;\;\;\;{\sum_{k\in {\cal U}_v}a_{vk}^q}\! \leqslant1 ,\forall v \in {\cal V} , \forall q \in {\cal Q}},\\
&{\;\;\;\;\;\;\;\;\;\;\;\;\;\;\;\;{{\cal U}_v \subset {\cal L}_v,\forall v \in {\cal V}}},\\
&{\;\;\;\;\;\;\;\;\;\;\;\;\;\;\;\;{P\left(E\left(\boldsymbol{\hat{\Psi}}_{vk},\bm{a}_{vk}\right) \geqslant \xi\right)\geqslant \epsilon ,\boldsymbol{\hat{\Psi}}_{vk} \subset \boldsymbol{\Psi}_{vk},\forall k \in {\cal U}_v}},
\end{align}
\end{subequations}
where ${\cal U}_v$ is the set of users associated with server $v$ and ${\cal L}_v$ is the set of users located in the service area of server $v$. Constraints (\ref{eq:optimal_problem}a), (\ref{eq:optimal_problem}b), and (\ref{eq:optimal_problem}c) ensure that each server can allocate one RB to each associated user and an RB can only be occupied by one user for image semantic information transmission. Constraint (\ref{eq:optimal_problem}e) is the semantic reliability requirement of each user. Since constraint (\ref{eq:optimal_problem}e) is non-convex and the semantic information extraction depends on deep neural network models, the problem (\ref{eq:optimal_problem}) cannot be solved by traditional optimization algorithms in polynomial time. Furthermore, a single server cannot observe the global wireless communication environment and the information of users associated with other servers. Hence, the centralized reinforcement learning algorithms (e.g., DQN) can only minimize the transmission latency of the implemented server based on the partial observation. To solve problem (\ref{eq:optimal_problem}) that aims to minimize the sum of the average transmission latency of all users, we introduce a multi-agent reinforcement learning algorithm that enables all servers cooperatively optimize the resource allocation of the considered semantic communication network.
 
\section{Value Decomposition based Entropy-Maximized Multi-Agent RL Method}
 To effectively solve problem (\ref{eq:optimal_problem}), we introduce a value decomposition based \cite{VDN} entropy-maximized multi-agent RL (VD-ERL) algorithm to minimize the average transmission latency of all servers instead of individual server. We first introduce the components of the proposed VD-ERL method. Then, we introduce the training procedure of the proposed VD-ERL method.

\subsection{Components of VD-ERL Method}
In this section, we introduce the fundamental components of the proposed VD-ERL method as follows:
\begin{itemize}
\item{\emph{Agent}}: The agents are the servers that determine the RB allocation and the set of semantic triples that need to transmit to its associated users. 
\item{\emph{States}}: The state is defined as $\bm{s}=\left[\bm{s}_1,\ldots,\bm{s}_v,\ldots,\bm{s}_V\right]$ where $\bm{s}_v=\left[\bm{\gamma}_v,\bm{\beta}_v\right]$ represents the partial state of server $v$. $\bm{\gamma}_v=\left[\gamma_v^1,\ldots,\gamma_v^Q\right]$ is the vector of available RBs where $\gamma_v^q=0$ represents that RB $q$ has been allocated, and $\gamma_v^q=1$, otherwise. $\bm{\beta}_v=\left[\bm{\beta}_{v_1},\ldots,\bm{\beta}_{vk},\ldots,\bm{\beta}_{v_{\vert{\cal L}_v\vert}}\right]$ is the semantic triple score matrix of the users located in the coverage of server $v$ and is used to evaluate the semantic reliability, where $\vert{\cal L}_v\vert$ is the number of users in the service area of server $v$ and $\bm{\beta}_{vk}=\left[\beta\left(\bm{\psi}^1_{vk}\right),\ldots,\beta\left(\bm{\psi}^n_{vk}\right),\ldots,\beta\left(\bm{\psi}^{N_{vk}}_{vk}\right)\right]$ is the vector of scores of all semantic triples in semantic information $\bm{\Psi}_{vk}$. The score of each semantic triple $\bm{\psi}_{vk}^n$ can be given as
\begin{equation}
\beta\left(\bm{\psi}_{vk}^n\right)=\frac{\exp\left(\mu\left(e^n_{vk,i}\right)\mu\left(l_{vk,ij}^n\right)\mu\left(e^n_{vk,j}\right)\right)}{\sum_{n=1}^{N_{vk}} \exp\left(\mu\left(e^n_{vk,i}\right)\mu\left(l^n_{vk,ij}\right)\mu\left(e^n_{vk,j}\right)\right)},
\end{equation}
where $\mu\left(e^n_{vk,i}\right)$ is the probability of object $e^n_{vk,i}$ being detected from image $G_{vk}$ and $\mu\left(l_{vk,ij}^n\right)$ is the conditional probability of relationship $l_{vk,ij}^n$ being deduced given objects $e^n_{vk,i}$ and $e^n_{vk,j}$. In (9), $\mu\left(e^n_{vk,i}\right)$ and $\mu\left(l_{vk,ij}^n\right)$ can be obtained by a scene graph generation model \cite{TDE}. The score of each semantic triple $\bm{\psi}_{vk}^n$ represents the probability of extracting triple $\bm{\psi}_{vk}^n$ from the original image and will be used to the selection of the partial semantic information to be transmitted. In particular, the semantic triple that has a high score can contribute more to the semantic information. Here, we note that, each server $v$ can only observe its partial state $\bm{s}_v$.

\item{\emph{Actions}}: Each action $\bm{\alpha}_{v}$ of server $v$ is the RB allocation, which is given by:
\begin{equation}
\bm{\alpha}_v=\left[\bm{a}_{v1},\ldots,\bm{a}_{vk},\ldots,\bm{a}_{v{\vert {\cal L}_v}\vert}\right],
\end{equation}
where $\bm{a}_{vk}$ representing RB allocation vector is the variable of problem (\ref{eq:optimal_problem}). Then, the vector of all distributed servers' actions is $\bm{\alpha}=\left[\bm{\alpha}_1,\ldots,\bm{\alpha}_v,\ldots,\bm{\alpha}_V\right]$.

\item{\emph{Policy}}: The policy is the conditional probability of each agent choosing an action $\bm{\alpha}_v$ in a given partial state $\bm{s}_v$. The policy is implemented by the DNN with parameter $\bm{\phi}_v$, which establishes the relation between the semantic triple scores, the ISS, and the transmission latency of each user. Then, the conditional probability of each agent taking action $\bm{\alpha}_v$ in a given partial state $\bm{s}_v$ can be expressed as $\bm{\pi}_{\bm{\phi}_v}\left(\bm{\alpha}_v\mid \bm{s}_v\right)$. To improve the action exploration, the policy networks are trained to maximize not only the expected reward, but also the entropy of actions ${\cal H}\left(\bm{\pi}_{\bm{\phi}_v}\left(\bm{\alpha}_v\mid \bm{s}_v\right)\right)$ which drives the agent to choose actions more randomly.

\item{\emph{Reward}}: The reward of each server is used to capture the benefits of a selected action in terms of semantic reliability and transmission latency. To calculate the reward of each server $v$, we first need to determine the partial semantic information $\bm{\hat{\Psi}}_{vk}$ that will be transmitted to user $k$. In particular, based on the state $\bm{s}_v$ and action $\bm{\alpha}_v$, we can sort the semantic triples $\bm{\Psi}_{vk}$ according to the score vector $\bm{\beta}_{vk}$. In particular, the sorted semantic triple vector is $\left[\bm{\hat{\psi}}^1_{vk},\ldots,\bm{\hat{\psi}}^{N_{vk}}_{vk}\right]$ where $\bm{\hat{\psi}}^1_{vk}$ is the triple with the highest score while $\bm{\hat{\psi}}^{N_{vk}}_{vk}$ is the triple with the lowest score. Given this sorted triple vector, we use an iterative algorithm to select several triples to satisfy constraint (8e) while minimizing the transmission time. The iterative algorithm used to determine the selected triples $\left[\bm{\hat{\psi}}^1_{vk},\ldots,\bm{\hat{\psi}}^n_{vk}\right]$ to generate semantic information is summarized in \textbf{Algorithm 1}.

Then, the reward of each user $k$ associated with server $v$ is given as
\begin{equation}\label{eq:userreward}
r_{vk}\left(\bm{s}_v,\bm{a}_{vk}\right)=\sum_{q=1}^Qa_{vk}^q\left[\mathbbm{1}_{\left\{E\left(\bm{\hat{\Psi}}_{vk},\bm{a}_{vk}\right)\geqslant \xi\right\}}\cdot\eta-T\left(\bm{\hat{\Psi}}_{vk},\bm{a}_{vk}\right)\right]
\end{equation}
 where $\eta$ is a constant bias, $T\left(\bm{\hat{\Psi}}_{vk},\bm{a}_{vk}\right)$ is the transmission latency, and $\mathbbm{1}_{\left\{E\left(\bm{\hat{\Psi}}_{vk},\bm{a}_{vk}\right)\geqslant {\xi}\right\}}$ is a function that indicates if the received semantic information $\bm{\hat{\Psi}}_{vk}$ satisfies the semantic reliability requirement defined in constraint (8e).
 
 Since servers allocate RB resources to users in a distributed manner and each server does not know the RB allocation schemes of other servers, several servers may allocate their RB to one user and this user can use the RB of only one server thus wasting the RB of other servers. To improve RB usage, we add a negative penalty $\rho$ to the reward function. In particular, the total reward of all servers is given as
 \begin{equation}\label{eq:totalreward}
 \begin{aligned}
r\left(\bm{s},\bm{\alpha}\right)&=\sum_{v=1}^V{r_v\left(\bm{s}_v,\bm{\alpha}_v\right)}\\
&=\sum_{v=1}^V{\sum_{k\in{\cal U}_v}}\left[\mathbbm{1}_{\left\{\sum_{\zeta\neq v,\zeta\in{\cal V}}\sum_{q=1}^Q\bm{a}_{vk}^q=0\right\}}r_{vk}\left(\bm{s}_v,\bm{a}_{vk}\right)+\left(1-\mathbbm{1}_{\left\{\sum_{\zeta\neq v,\zeta\in{\cal V}}\sum_{q=1}^Q\bm{a}_{vk}^q=0\right\}}\right)\rho\right],
\end{aligned}
\end{equation}
where $r_v\left(\bm{s}_v,\bm{\alpha}_v\right)$ is the reward of server $v$ and $\mathbbm{1}_{\left\{\sum_{\zeta\neq v,\zeta\in{\cal V}}\sum_{q=1}^Q\bm{a}_{vk}^q=0\right\}}$ is a function that indicates whether user $k$ is served by other servers. From (\ref{eq:totalreward}), we see that, when an RB is underutilized, the reward will be $\rho$.
 
\begin{algorithm}[t]
\small
\caption{Semantic triples selection algorithm.}
\begin{algorithmic}[1]
\STATE \textbf{Input:} The distribution of semantic triple scores $\bm{\beta}_{vk}$, the number of semantic triples $N_{vk}$, and the minimum semantic similarity $\xi$.
\STATE \textbf{Initialize:} Sorting the semantic triples in the descending order in $\bm{\Psi}_{vk}$ by $\bm{\beta}_{vk}$.  
\FOR {$n = 1 \to N_{vk}$}
\STATE Select $n$ triples with highest score $\bm{\hat{\Psi}}_{vk}=\left[\bm{\hat{\psi}}^1_{vk},\ldots,\bm{\hat{\psi}}^n_{vk}\right]$.
\STATE Estimate semantic similarity by semantic triple scores $\widetilde{E}\left(\bm{\hat{\Psi}}_{vk}\right)=\sum_{i=1}^{n}\bm{\beta}\left(\bm{\hat{\psi}}^i_{vk}\right)$.
\IF{$\widetilde{E}\left(\bm{\hat{\Psi}}_{vk}\right) \geqslant \xi$}  
\STATE \textbf{end for}
\ENDIF
\ENDFOR
\STATE \textbf{Output:} Selected semantic triples ${\bm{\hat{\Psi}}_{vk}}$.
\end{algorithmic}
\label{algorithm_1}
\end{algorithm}

\item{\emph{Individual Q value function}}: The individual Q value function of each server $v$ is defined as $Q_{\bm{\theta_v}}\left(\bm{s}_v,\bm{\alpha}_{v}\right)$, which is used to estimate the expected reward under a given partial state $\bm{s}_v$ of server $v$ and a selected action $\bm{\alpha}_v$. Each server $v$ uses a DNN with parameter $\bm{\theta}_v$ to approximate the individual Q value function. Since each server can observe only the state of the users located in its service area, each server will transmit its individual Q value to other servers for the estimation of global Q value function, which will be explained in the next bullet.
\item{\emph{Global Q value function}}: The global Q value function is defined as $Q_{tot}\left(\bm{s},\bm{\alpha}\right)$, which is used to estimate the total expected reward of all distributed servers. For the proposed VD-ERL algorithm, we assume that the global Q value of all servers is equal to the sum of the individual Q value of each servers, which is given by \cite{VDN}
\begin{equation}
Q_{tot}\left(\bm{s},\bm{\alpha}\right)=\sum_{v=1}^V Q_{\bm{\theta}_v}\left(\bm{s}_v,\bm{\alpha}_v\right).    
\end{equation}

The goal of each server $v$ is to cooperatively maximize the total expected reward, i.e., maximize the global Q value by training its policy network. After training, each server can find the optimal policy based on the global Q value function so as to minimize the sum of the transmission latency of all users while satisfying their semantic reliability requirements.
\end{itemize}
\begin{figure}[t]\centering
\includegraphics[width=16cm]{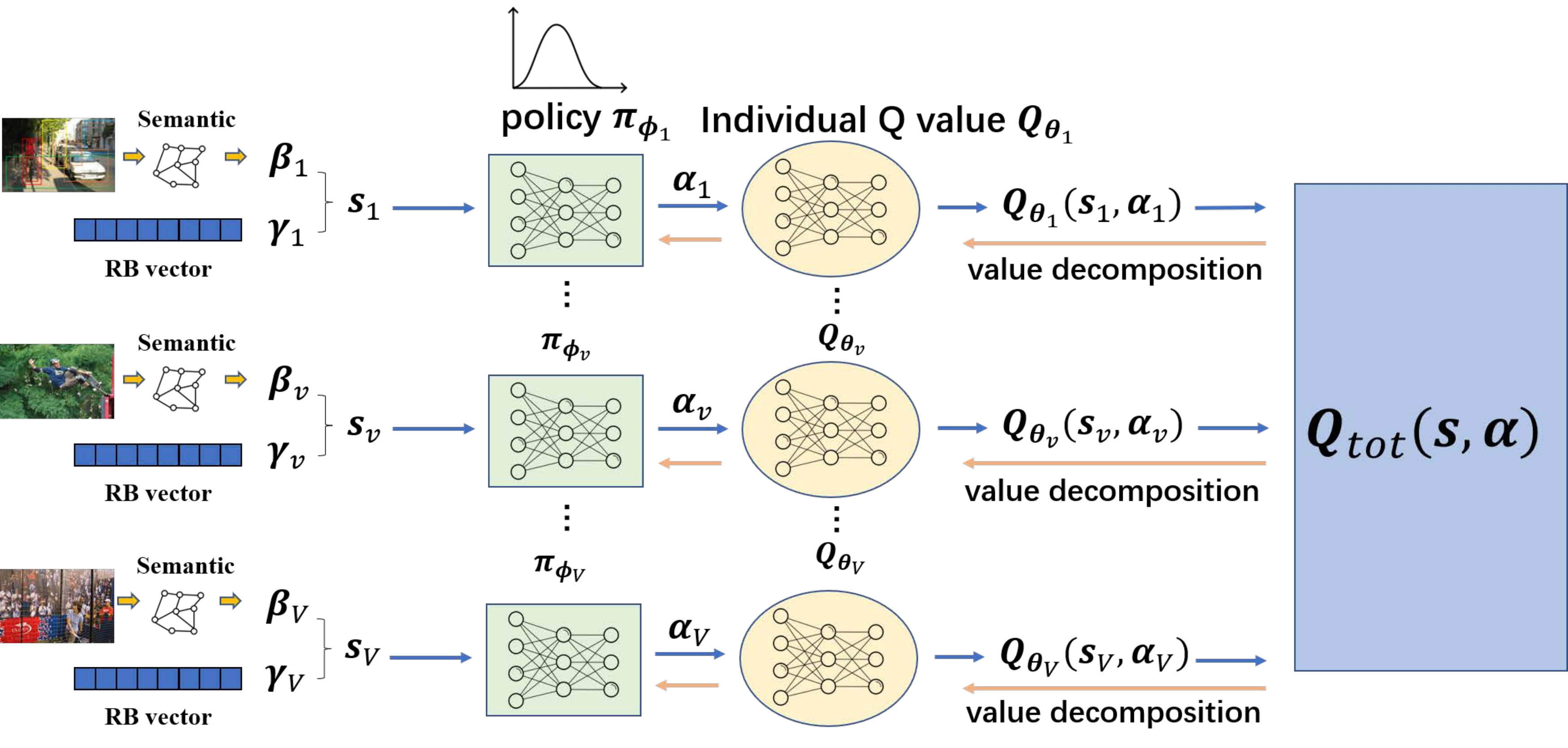}
\caption{The training process of the proposed VD-ERL algorithm.}
\label{fig4}
\end{figure}

\subsection{VD-ERL Algorithm for Semantic Oriented Resource Allocation}
 Next, we introduce how the servers use the proposed VD-ERL algorithm to cooperatively minimize the sum of the average semantic information transmission latency. At first, each agent first collects local information that includes partial states $\bm{s}_v$ and actions $\bm{\alpha}_v$. Then, each agent transmits its local information to other agents to calculate its server reward $r_v\left(\bm{s}_v,\bm{\alpha}_v\right)$ and total reward $r\left(\bm{s},\bm{\alpha}\right)=\sum_{v=1}^Vr_v\left(\bm{s}_v,\bm{\alpha}_v\right)$ of all servers. Finally, as shown in Fig.~\ref{fig4}, based on the total reward and global Q value function, each agent updates the its policy network and individual Q value function. In particular, each agent first collects a set of trajectories ${\cal D}_v =\left\{\bm{\tau}_v^1,\ldots,\bm{\tau}_v^d,\ldots,\bm{\tau}_v^D\right\}$ with $\bm{\tau}_v^d=\left[\bm{\alpha}_v^d,\bm{s}_v^d,r_v^d\right]$ based on the current policy $\bm{\pi}_{\bm{\phi}_v}\left(\bm{s}_v^d,\bm{\alpha}_v^d\right)$. Then, each agent samples a batch size of trajectories from ${\cal D}_v$ and calculate total reward and global Q value to train individual Q value function $Q_{{\bm{\theta}}_v}$ and policy network $\bm{\pi}_{\bm{\phi}_v}$. Finally, each server samples action $\bm{\alpha}_v$ based on updated policy network $\bm{\pi}_{\bm{\phi}_v}$ under given state $\bm{s}_v$ to collect new trajectories for next iteration. The loss function of global Q value function $Q_{tot}\left(\bm{s},\bm{\alpha}\right)$ is defined as follows 
\begin{equation}
\begin{aligned}
J\left(\bm{\theta}_1,\ldots,\bm{\theta}_V\right)=\mathbb{E}_{\bm{\tau}^d\in {\cal D}}\left[Q_{tot}\left(\bm{s},\bm{\alpha}\right)-r\left(\bm{s},\bm{\alpha}\right)-\max\limits_{\bm{\alpha}'}Q_{tot}\left(\bm{s}',\bm{\alpha}'\right)\right]^2,
\end{aligned}
\end{equation}
where $\max\limits_{\bm{\alpha}'}Q_{tot}\left(\bm{s}',\bm{\alpha}'\right)$ is the maximal global Q value of next state $\bm{s}'$. The global Q value monotonically increases as each individual Q value increases, i.e., an action of an agent with a high individual Q value is also valuable for entire wireless networks. Hence, the goal that each server trains its individual Q value function is to maximize the global Q value. The individual Q value function $Q_{\bm{\theta}_v}\left(\bm{s}_v,\bm{\alpha}_v\right)$ of each server can be updated using a gradient descent method as follows:
\begin{equation}
\bm{\theta}_v\leftarrow\bm{\theta}_v-\lambda_{\bm{\theta}_v}\nabla_{\bm{\theta}_v} J\left(\bm{\theta}_1,\ldots,\bm{\theta}_V\right),
\end{equation}
where $\lambda_{\bm{\theta}_v}$ is the updating rate and $\nabla_{\bm{\theta}_v} J\left(\bm{\theta}_1,\ldots,\bm{\theta}_V\right)$ is the gradient of the global Q value function, which is given by
\begin{equation}
\begin{split}
\nabla_{\bm{\theta}_v} J\left(\bm{\theta}_1,\ldots,\bm{\theta}_V\right)&=\nabla_{\bm{\theta}_v}\left[Q_{tot}\left(\bm{s}^d,\bm{\alpha}^d\right)-r\left(\bm{s}^d,\bm{\alpha}^d\right)-\max\limits_{\bm{\alpha}'}Q_{tot}\left(\bm{s}',\bm{\alpha}'\right)\right]^2\\
&=2\Delta Q_{tot} \nabla_{\bm{\theta}_v}Q_{tot}\left(\bm{s}^d,\bm{\alpha}^d\right)\cdot\nabla_{\bm{\theta}_v}Q_{\bm{\theta}_v}\left(\bm{s}_v^d,\bm{\alpha}_v^d\right),
\end{split}
\end{equation}
where $\Delta Q_{tot}=Q_{tot}\left(\bm{s}^d,\bm{\alpha}^d\right)-r\left(\bm{s}^d,\bm{\alpha}^d\right)-\max\limits_{\bm{a}'}Q_{tot}\left(\bm{s}',\bm{\alpha}'\right)$. Combined with entropy-maximization RL \cite{SAC}, the objective of the policy $\bm{\pi}_{\bm{\phi}_v}$ is the weighted sum of the expected reward and the entropy of actions. Hence, the loss function of a policy network is given as 
\begin{equation}
\begin{aligned}
J_{\bm{\pi}}\left(\bm{\phi}_v\right)&=-\delta{\cal H}\left[\bm{\pi}_{\bm{\phi}_v}\left(\bm{\alpha}_v\mid\bm{s}_v^d\right)\right]-\mathbb{E}_{{\bm{s}_v^d\in \cal{D}},{\bm{\alpha}_v\in \bm{\pi}}_{\bm{\phi}_v}}\left[r_v\left(\bm{s}_v^d,\bm{\alpha}_v\right)\right]\\
&=\mathbb{E}_{{\bm{s}_v^d\in \cal{D}},{\bm{\alpha}_v\in \bm{\pi}}_{\bm{\phi}_v}}\left[\delta log{\bm{\pi}}_{\bm{\phi}_v}\left(\bm{\alpha}_v \mid \bm{s}_v^d\right)-Q_{\bm{\theta}_v}\left(\bm{s}_v^d, \bm{\alpha}_v\right)\right]\\
&=\mathbb{E}_{\bm{s}_v^d\in \cal{D}}\left[D_{KL}\left[\delta \bm{\pi_{\bm{\phi}_v}}\left(\bm{\alpha}_v \mid \bm{s}_v^d\right)\Vert \exp\left({Q_{\bm{\theta}_v}\left(\bm{s}_v^d, \bm{\alpha}_v\right)}\right)\right]\right],\\
\end{aligned}
\end{equation}
where $\delta$ is the temperature parameter to adjust the weight of the entropy term and the policy of each server will be more randomly as $\delta$ increases. From (17), we can see that the objective of the policy network is equivalent to minimizing the Kullback-Leibler (KL) divergence between the conditional probability distribution $\bm{\pi}_{\bm{\phi}_v}\left(\bm{\alpha}_v\mid\bm{s}_v^d\right)$ and the corresponding trained individual Q value function $Q_{\bm{\theta}_v}\left(\bm{\alpha}_v\mid\bm{s}_v^d\right)$. Hence, the action achieving higher individual Q value will be assigned a higher selection probability to be chosen under given local observation. However, the introduced entropy-maximization enables other potential valuable actions with low selection probability to be taken properly. Therefore, the valuable action exploration ability of each server and the probability of finding an optimal RB allocation scheme are improved. Finally, the policy ${\bm{\pi}}_{\bm{\phi}_v}$ can be updated using gradient descent method as follows:
\begin{equation}
\bm{\phi}_v \leftarrow \bm{\phi}_v-\lambda_{\bm{\phi}_v}\nabla J_{\bm{\pi}}\left(\bm{\phi}_v\right), 
\end{equation}
where $\lambda_{\bm{\phi}_v}$ is the learning rate. The specific training procedure of the proposed VD-ERL algorithm is summarized in \textbf{Algorithm 2}.

\subsection{Complexity and Convergence of the Proposed Algorithm}
In this section, we analyze the complexity and convergence of the proposed VD-ERL algorithm for semantic-oriented RB allocation. The complexity of the VD-ERL algorithm lies in semantic triple selection and determining the resource allocation of each server. First, from \textbf{Algorithm 1}, the complexity of semantic triple selection of user $k$ is ${\cal O}\left(N_{vk}\right)$. Hence, the complexity of semantic triple selection of all users is ${\cal O}\left(\sum_{v\in{\cal V}}\sum_{k=1}^{\vert{\cal U}_v\vert}N_{vk}\right)={\cal O}\left(N\right)$. Then, we explain the complexity of training policy and individual Q value networks of each server, which are two fully connected networks that consist of an input layer, hidden layers, and an output layer. Hence, the time-complexity of training networks of each server depends on the the number of neurons in each layer \cite{VLC}. The time-complexity of each network is ${\cal O}\left(\sum_{i=1}^{I-1} w_iw_{i+1}+\left({\vert{\cal L}_v\vert}N_{vk}+Q\right)w_1+\vert{\cal A}_v\vert w_I\right)$, where $w_i$ is the number of neurons in the hidden layer $i$, $I$ is the number of hidden layers, ${\vert{\cal L}_v\vert}N_{vk}+Q$ and $\vert{\cal A}_v\vert$ represent the dimension of input and output layer, respectively. The proposed algorithm can be trained offline. Therefore, \textbf{Algorithm 1} and \textbf{Algorithm 2} are executed with complexity ${\cal O}\left(N+\sum_{i=1}^{I-1} w_iw_{i+1}+\left({\vert{\cal L}_v\vert}N_{vk}+Q\right)w_1+\vert{\cal A}_v\vert w_I\right)$ in the training stage. After training, we only need to implement \textbf{Algorithm 1} for RB allocation with complexity ${\cal O}\left(N\right)$.

Next, using the result of \cite[Theorem 1]{SAC} , we can prove that the proposed VD-ERL algorithm 
is guaranteed to converge to a locally optimal solution of problem (\ref{eq:optimal_problem}), as shown in the following lemma.

\begin{lemma} \emph{The proposed VD-ERL algorithm is guaranteed to converge if the following conditions are satisfied:
1) Individual Q value function $Q_{\bm{\theta_v}}\left(\bm{s}_v,\bm{\alpha}_{v}\right)$ is bounded.
2) $Q^{\bm{\pi}_{\bm{\phi}_v}^{\textrm{N}}}_{\bm{\theta_v}}\left(\bm{s}_v,\bm{\alpha}_{v}\right)\geqslant Q^{\bm{\pi}_{\bm{\phi}_v}^{\textrm{O}}}_{\bm{\theta_v}}\left(\bm{s}_v,\bm{\alpha}_{v}\right)$ holds for any state $\bm{s}_v$ and action $\bm{\alpha}_{v}$, where ${\bm{\pi}_{\bm{\phi}_v}^{\textrm{N}}}$ is the optimized policy based on (17) with current individual Q value function in each iteration.}
\end{lemma}

\emph{Proof:} Next, we prove that the proposed VD-ERL algorithm satisfies these two conditions. Since the number of actions $\bm{\alpha}$ in the proposed VD-ERL algorithm is finite, the global Q value function $Q_{tot}\left(\bm{s},\bm{\alpha}\right)$ can be proved to be bounded using \cite[Lemma 1]{SAC}. Hence, condition 1) is satisfied. For condition 2), from (17), the new policy ${\bm{\pi}_{\bm{\phi}_v}^{\textrm{N}}}$ satisfies the following inequality equation for any old policy ${\bm{\pi}_{\bm{\phi}_v}^{\textrm{O}}}$:
\begin{equation}
D_{KL}\left[{\bm{\pi}_{\bm{\phi}_v}^{\textrm{N}}}\left(\bm{\alpha}_v \mid \bm{s}_v\right)\Vert \exp\left({Q_{\bm{\theta}_v}\left(\bm{s}_v, \bm{\alpha}_v\right)}\right)\right]\leqslant D_{KL}\left[ {\bm{\pi}_{\bm{\phi}_v}^{\textrm{O}}}\left(\bm{\alpha}_v \mid \bm{s}_v\right)\Vert \exp\left({Q_{\bm{\theta}_v}\left(\bm{s}_v, \bm{\alpha}_v\right)}\right)\right].
\end{equation}
Given (19), we can prove that the proposed method satisfies condition 2) using the result of \cite[Lemma 2]{SAC}. 

\begin{table}\centering\footnotesize
\setlength{\belowcaptionskip}{0pt}
\setlength{\abovedisplayskip}{-15pt}
\setlength{\tabcolsep}{5mm}{
\newcommand{\tabincell}[2]{\begin{tabular}{@{}#1@{}}#2.0\end{tabular}}
\renewcommand\arraystretch{1}
\caption[table]{{System Parameters}}
\label{x}\centering
\begin{tabular}{|c|c|c|c|}
\hline
\!\textbf{Parameter}\! \!\!& \!\!\!\!\textbf{Value} &\! \textbf{Parameter} \!& \!\!\!\!\textbf{Value}\!\!\! \\
\hline
$Q$ & 8 & $W$ & 2 MHz \\
\hline
$V$ & 5 &  $U$ & 50  \\
\hline
$P$ & 1 W  & $N_0$ &-174 dBm/Hz \\
\hline
$\eta$ & 3 & $\rho$ & -3 \\
\hline
 $\epsilon$ &0.9 & $\xi$ & 0.5  \\
\hline
\end{tabular}}
\end{table}

\begin{algorithm}[t]
\small
\caption{VD-ERL algorithm for solving problem (\ref{eq:optimal_problem}).}
\begin{algorithmic}[2]
\STATE \textbf{Initialize:} Networks parameters $\left\{\bm{\theta}_1,\ldots,\bm{\theta}_V\right\},\left\{\bm{\phi}_1,\ldots,\bm{\phi}_V\right\}$ generated randomly, learning rate and update rate $\left\{\lambda_{\bm{\theta}_1},\ldots,\lambda_{\bm{\theta}_V}\right\},\left\{\lambda_{\bm{\phi}_1},\ldots,\lambda_{\bm{\phi}_V}\right\}$, and the number of iterations $N$.
\FOR {$i = 1 \to N$}
\FOR {each environment step}
\FOR {each agent}
\STATE Record local observation of environment state $\bm{s}_v$.
\STATE Select an action $\bm{\alpha}_v$ based on current policy $\bm{\pi}_{\bm{\phi}_v}$ 
\STATE Transmit the action $\bm{\alpha}_v$ and state $\bm{s}_v$ to other agents.
\STATE Calculate the server reward of each server and collect a series of trajectories ${\cal D}_v={\cal D}_v$ $\cup$ $\left\{\left(\bm{\alpha}_v,\bm{s}_v,r_v\left(\bm{s}_v,\bm{\alpha}_v\right)\right)\right\}$.
\ENDFOR
\ENDFOR
\FOR {each gradient step}
\STATE Calculate the total reward $r\left(\bm{s},\bm{\alpha}\right)$ and global Q value $Q_{tot}\left(\bm{s},\bm{\alpha}\right)$
\STATE Update $\left\{\bm{\theta}_1,\ldots,\bm{\theta}_V\right\}$ by (15).
\STATE Update $\left\{\lambda_{\bm{\phi}_1},\ldots,\lambda_{\bm{\phi}_V}\right\}$ by (18).
\ENDFOR
\ENDFOR
\end{algorithmic}
\label{algorithm_2}
\end{algorithm}

\section{Simulation Results and Analysis}   
For our simulations, we consider a circular wireless network area. In the considered network, five servers are deployed around the center to transmit image data to $U = 50$ uniformly distributed users. Other system parameters are listed in Table \ref{x}. We use the scene graph generation model in \cite{TDE} for semantic information extraction and the multimodal data embedded model in \cite{CLIP} for vectorization of semantic information and image. The visual genome (VG) \cite{VG} dataset is used to train the proposed algorithm. For comparison purposes, we consider three baselines of RB allocation methods: a) the random method, b) the independent deep Q learning method, and c) the value decomposition based deep Q learning network method. All experimental results are averaged over a large number of independent runs.

\begin{figure*}[tp]
\centering  
\includegraphics[width=13cm]{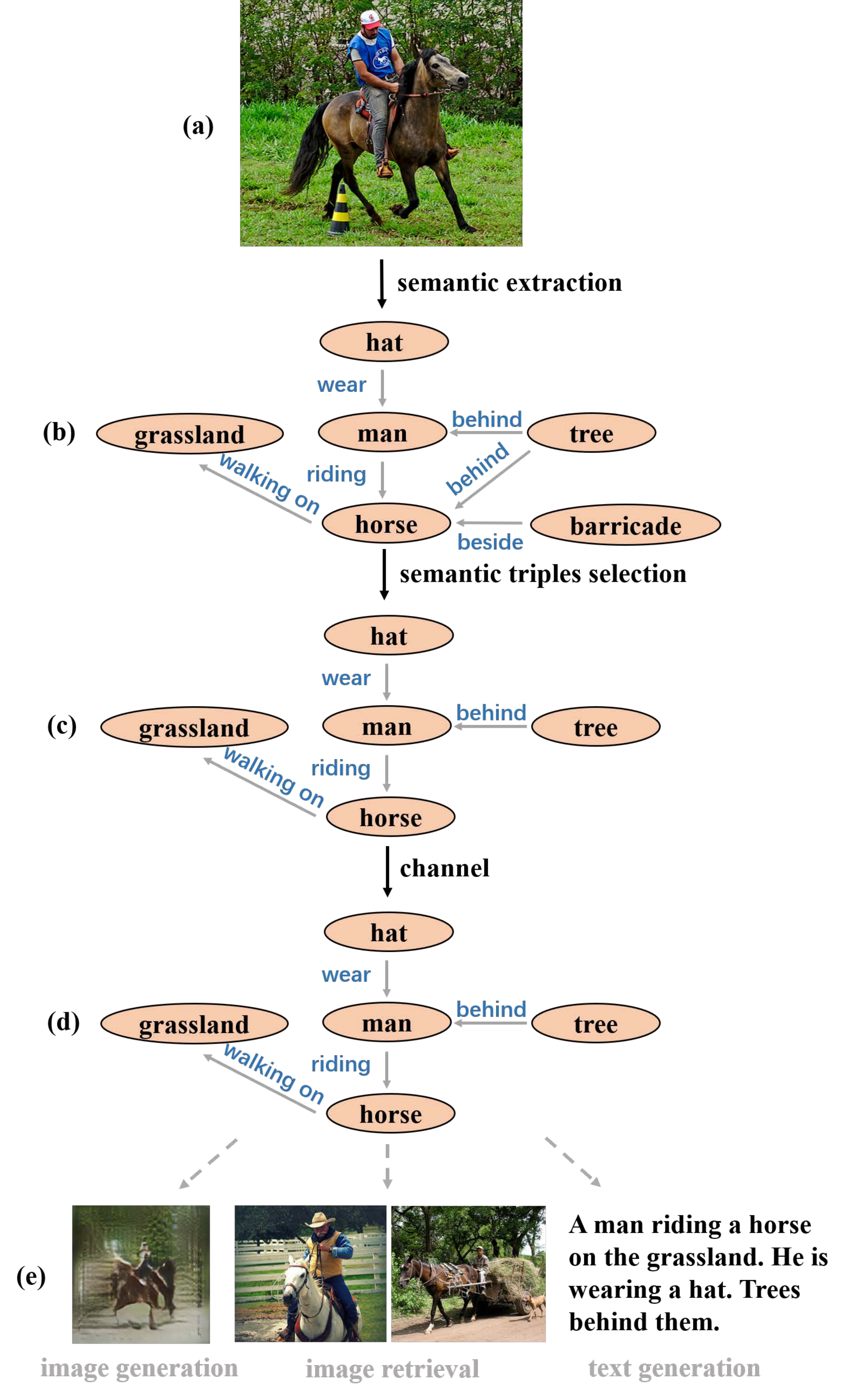}
\caption{An example of semantic communication system for image transmission.}
\label{fig5}
\end{figure*}

Figure \ref{fig5} shows an example of the image transmission using our designed semantic communication framework. In Fig. \ref{fig5}, the server needs to send an image, as shown in Fig. \ref{fig5}a), to a user. Then, the server uses a scene graph generation model to extract semantic information of this image, as shown in Fig. \ref{fig5}b). In Fig. \ref{fig5}b), we see that the objects and their corresponding relationships are extracted from the original image Fig. \ref{fig5}a). Given the user association and RB allocation schemes, the next step is to select triples to generate transmitted partial semantic information. Fig. \ref{fig5}c) shows the selected triples and generated partial semantic information. From Figs. \ref{fig5}b) and \ref{fig5}c), we can see that the triple ``\emph{barricade beside horse}'' and triple ``\emph{tree behind horse}'' are not selected to generate semantic information since these triples are trivial or redundant. This indicates that the proposed image semantic communication framework can find meaningless triples and do not use them for semantic information generation thus reducing the transmission delay by only transmitting partial important triples. Figure \ref{fig5}d) shows the semantic information received by the user. The user can use this semantic information to generate original image, retrieve images with similar semantic information, and generate a caption of the original image, as shown in Fig. \ref{fig5}e). 
In particular, Fig. \ref{fig5}e) shows the use of a generative adversarial network and the received semantic information to generate images that are similar to the original image in semantic level, which demonstrates that the extracted semantic information are meaningful enough for various applications.

\begin{figure}[t]\centering
\includegraphics[width=12cm]{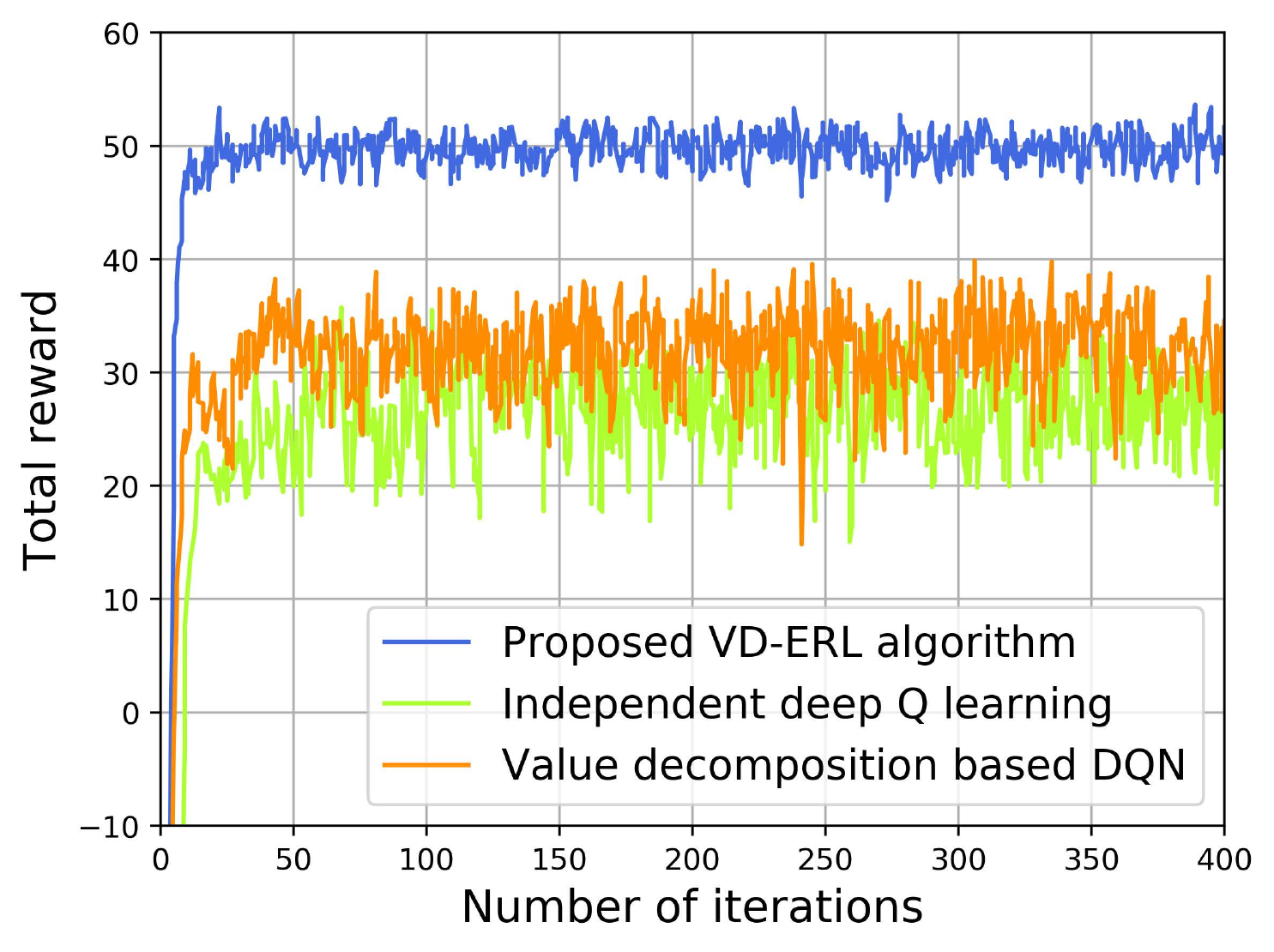}
\caption{The convergence of the proposed VD-ERL algorithm.}
\label{fig6}
\end{figure}

Figure 6 shows the convergence of the proposed VD-ERL algorithm. In Fig.~\ref{fig6}, we can see that the independent deep Q learning algorithm remains divergent after 100 iterations.  
Figure 6 also shows that, compared to the VD based DQN algorithm that converges after 60 iterations, the proposed VD-ERL algorithm converges after 30 iterations. 
This stems from the fact that the proposed VD-ERL algorithm utilizes a value network to evaluate and promote the policy network and hence, the minor change of value function can not change the action choose directly, which is indifferent in VD based DQN algorithm. From Fig.~\ref{fig6}, we can also observe that the proposed VD-ERL algorithm achieves 78.6\% and 42.9\% improvement in total reward compared to the independent deep Q learning algorithm and VD based DQN algorithm respectively. This is due to the fact that the proposed VD-ERL can optimize the action exploration by maximizing the policy entropy, which enables each server to find globally optimal RB allocation policy. 

\begin{figure}[t]
\centering  
\subfigure[Random RB allocation method as the number of users varies.]{
\centering
\includegraphics[width=7.8cm]{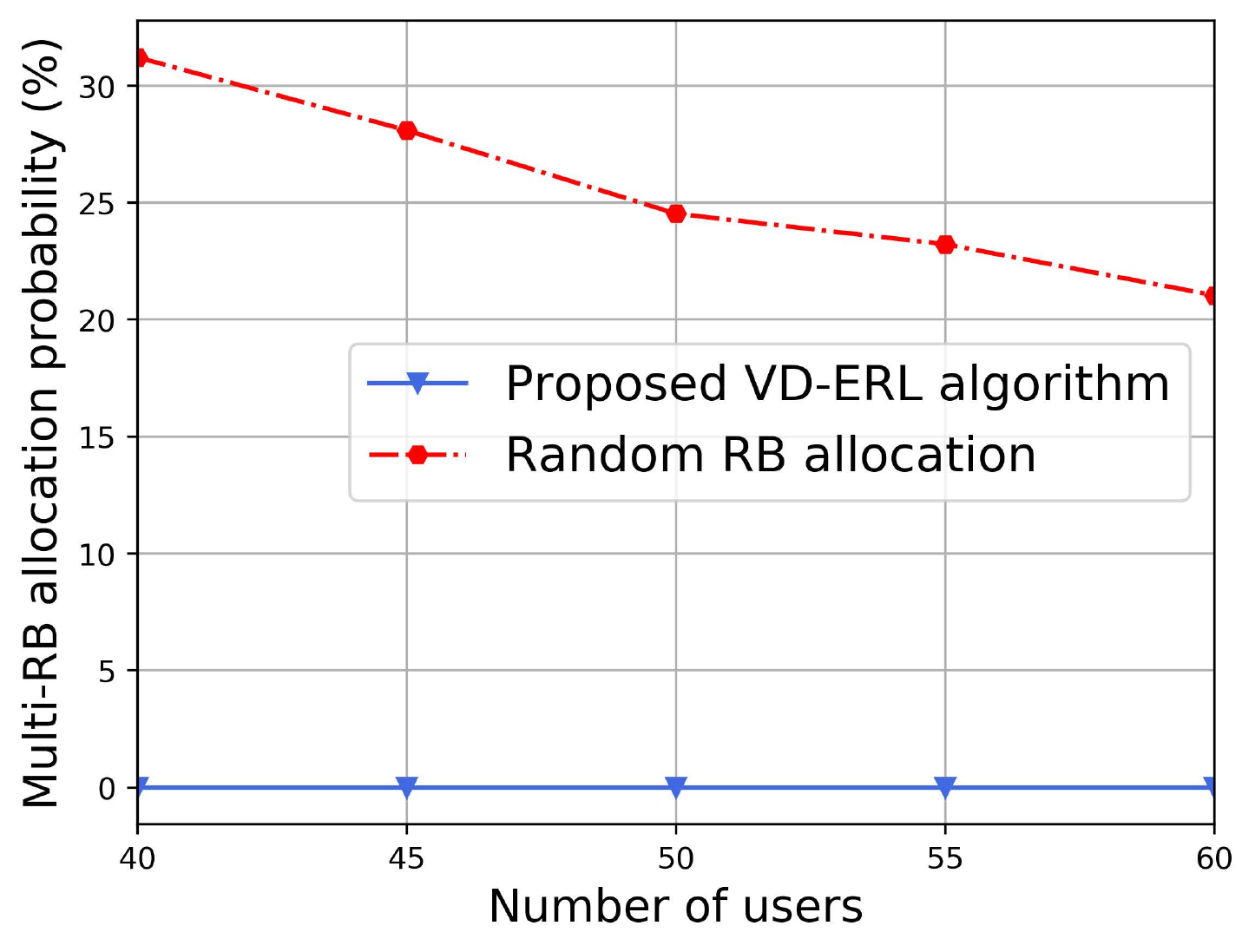}
}
\subfigure[Traditional multi-agent RL based methods as the number of users varies.]{
\centering
\includegraphics[width=7.8cm]{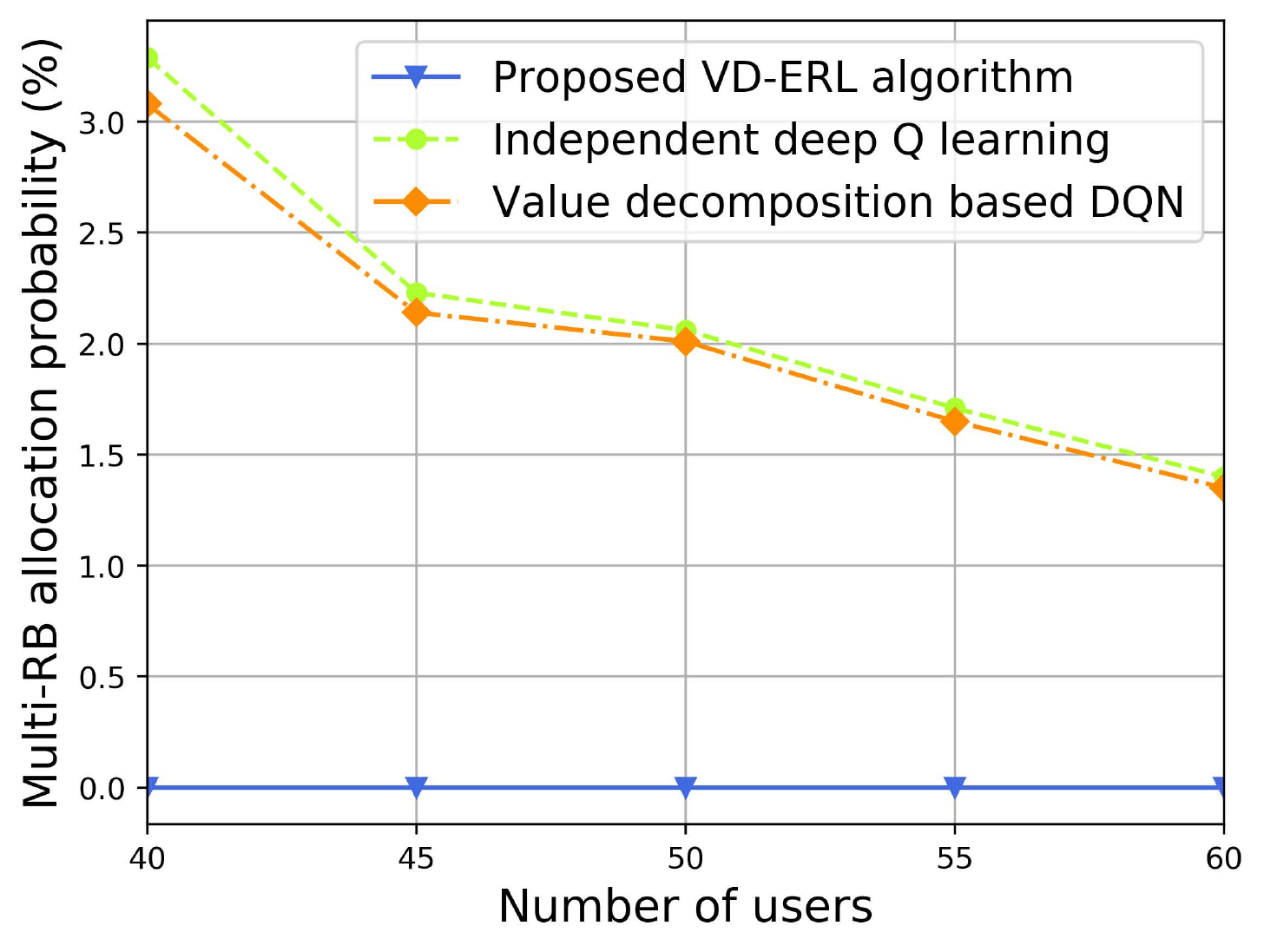}}
\subfigure[Random RB allocation method as the number of servers varies.]{
\centering
\includegraphics[width=7.8cm]{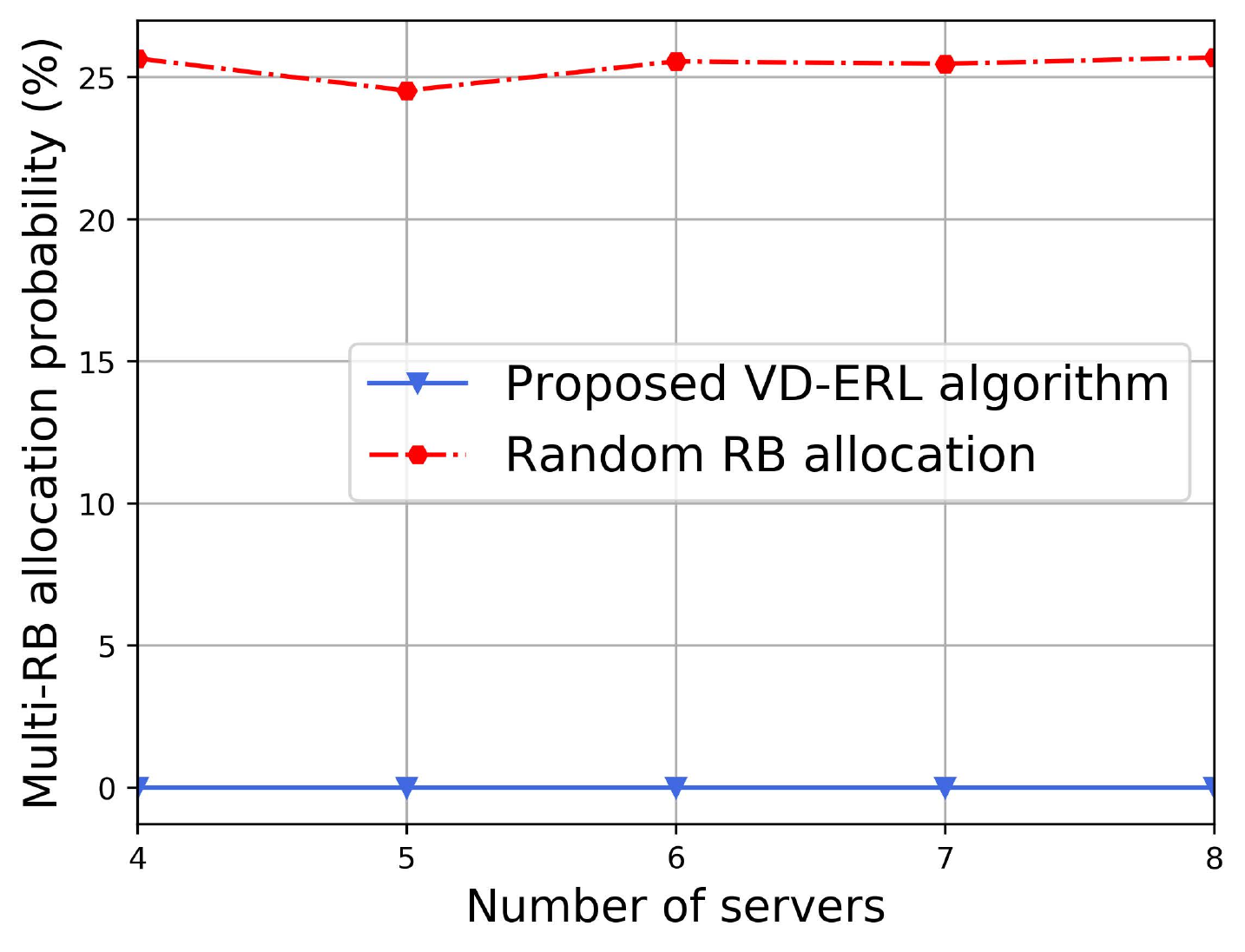}}
\subfigure[Traditional multi-agent RL based methods as the number of servers varies.]{
\centering
\includegraphics[width=7.8cm]{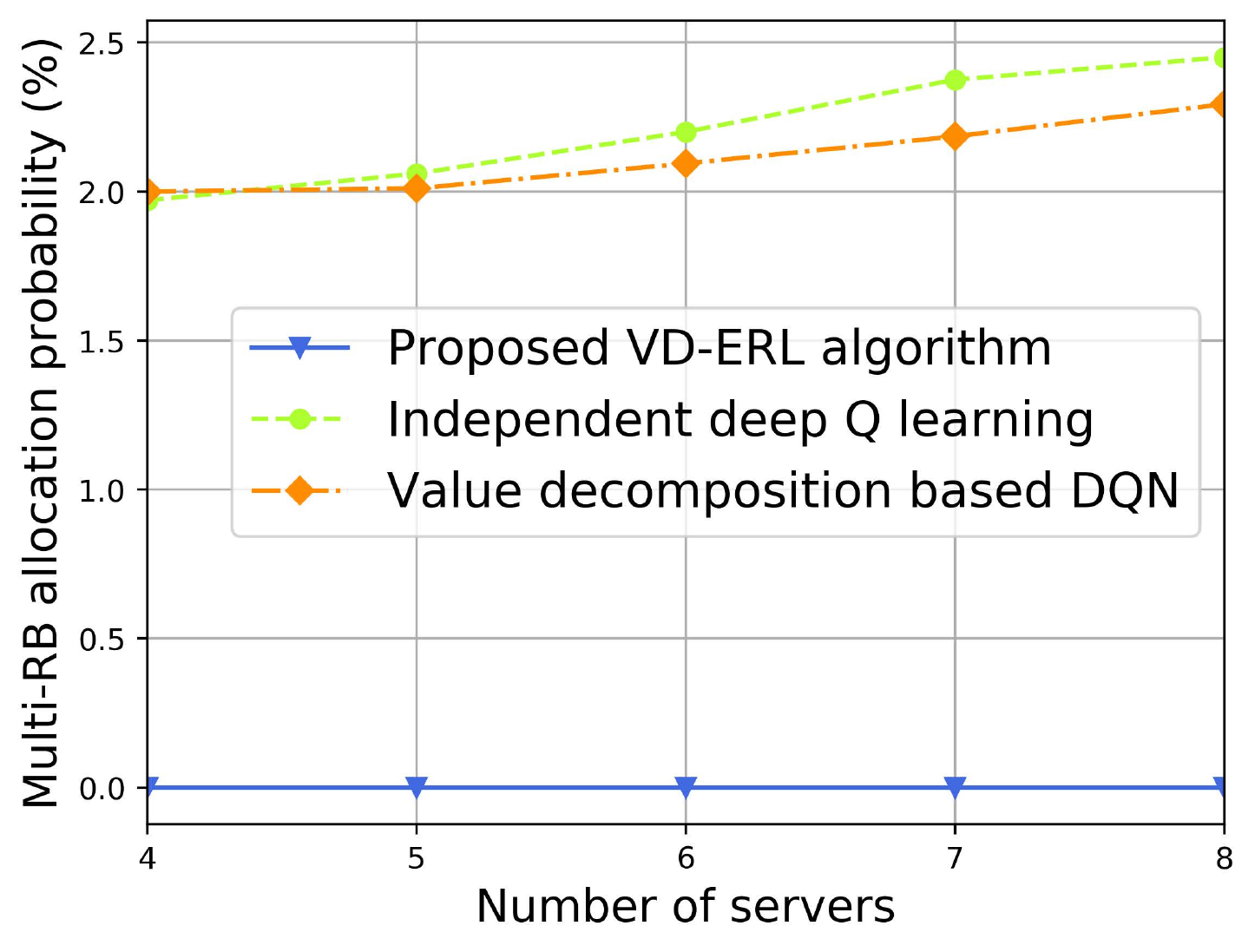}}
\caption{Multi-RB allocation probability of the proposed VD-ERL algorithm.}
\label{fig7}
\end{figure}

\begin{figure*}[t]
\centering  
\subfigure[Versus random method as the number of users varies.]{
\centering
\includegraphics[width=7.8cm]{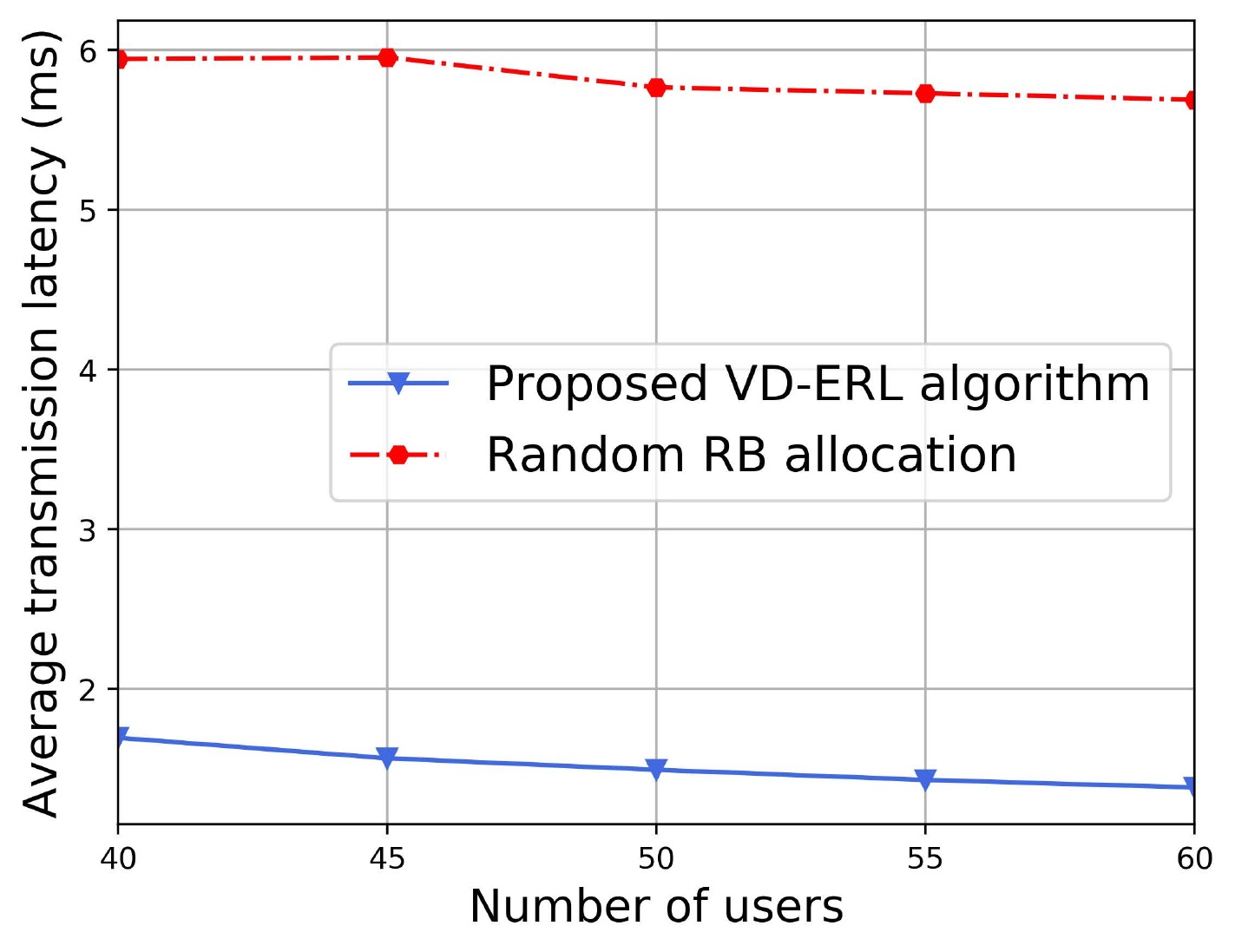}
}
\subfigure[Versus traditional multi-agent RL based methods as the number of users varies.]{
\centering
\includegraphics[width=7.8cm]{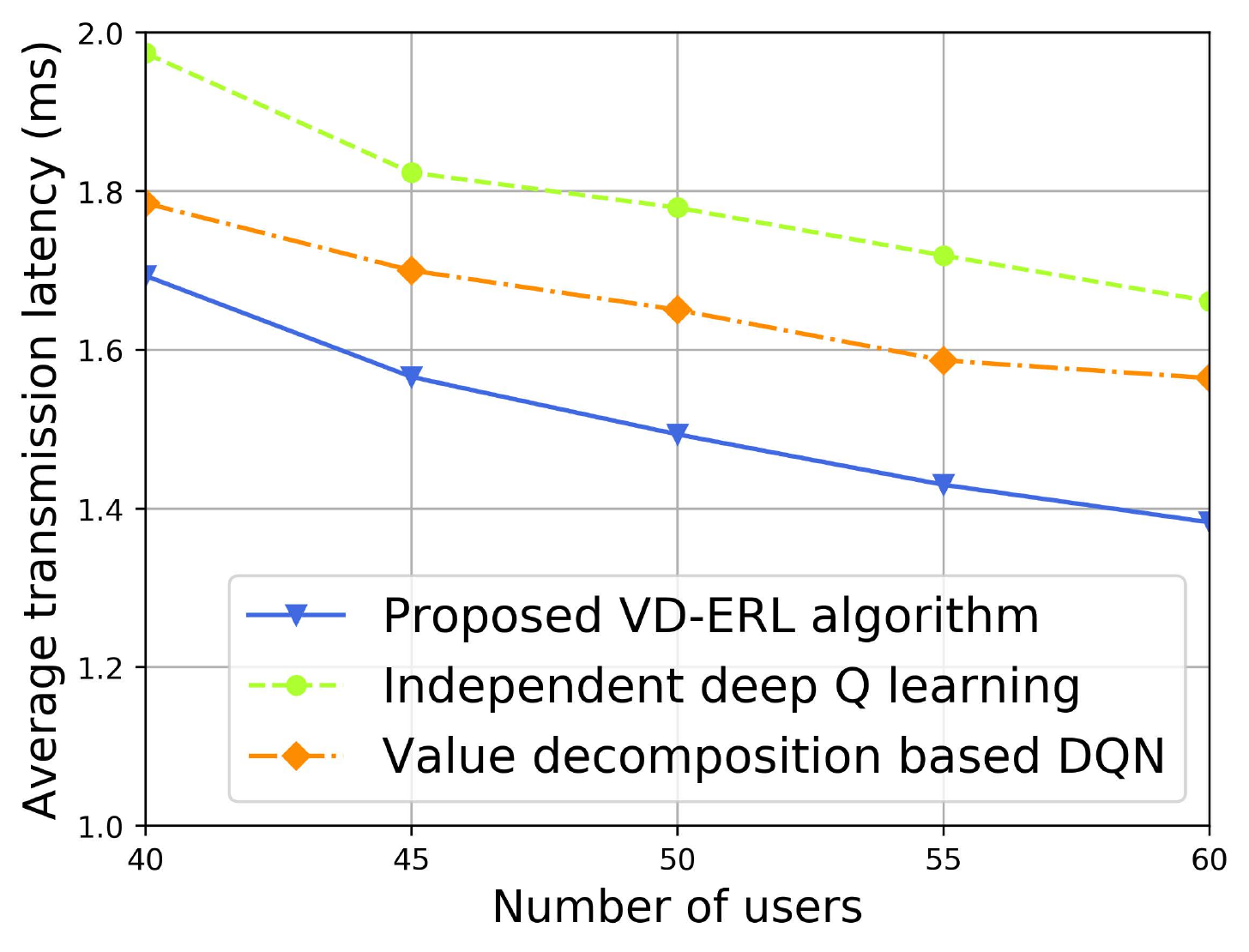}
}
\subfigure[Versus random method as the number of servers varies.]{
\centering
\includegraphics[width=7.8cm]{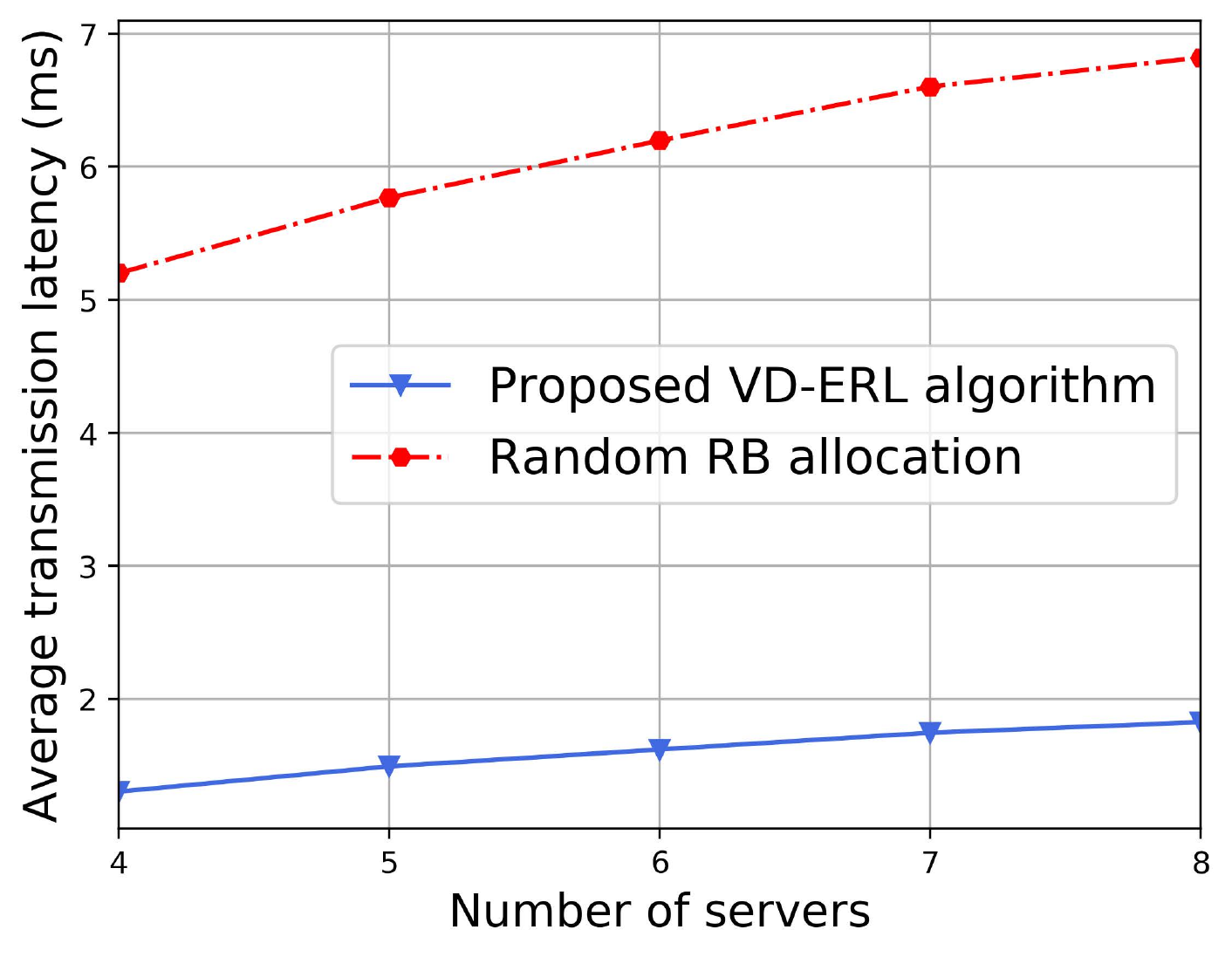}
}
\subfigure[Versus traditional multi-agent RL based methods as the number of servers varies.]{
\centering
\includegraphics[width=7.8cm]{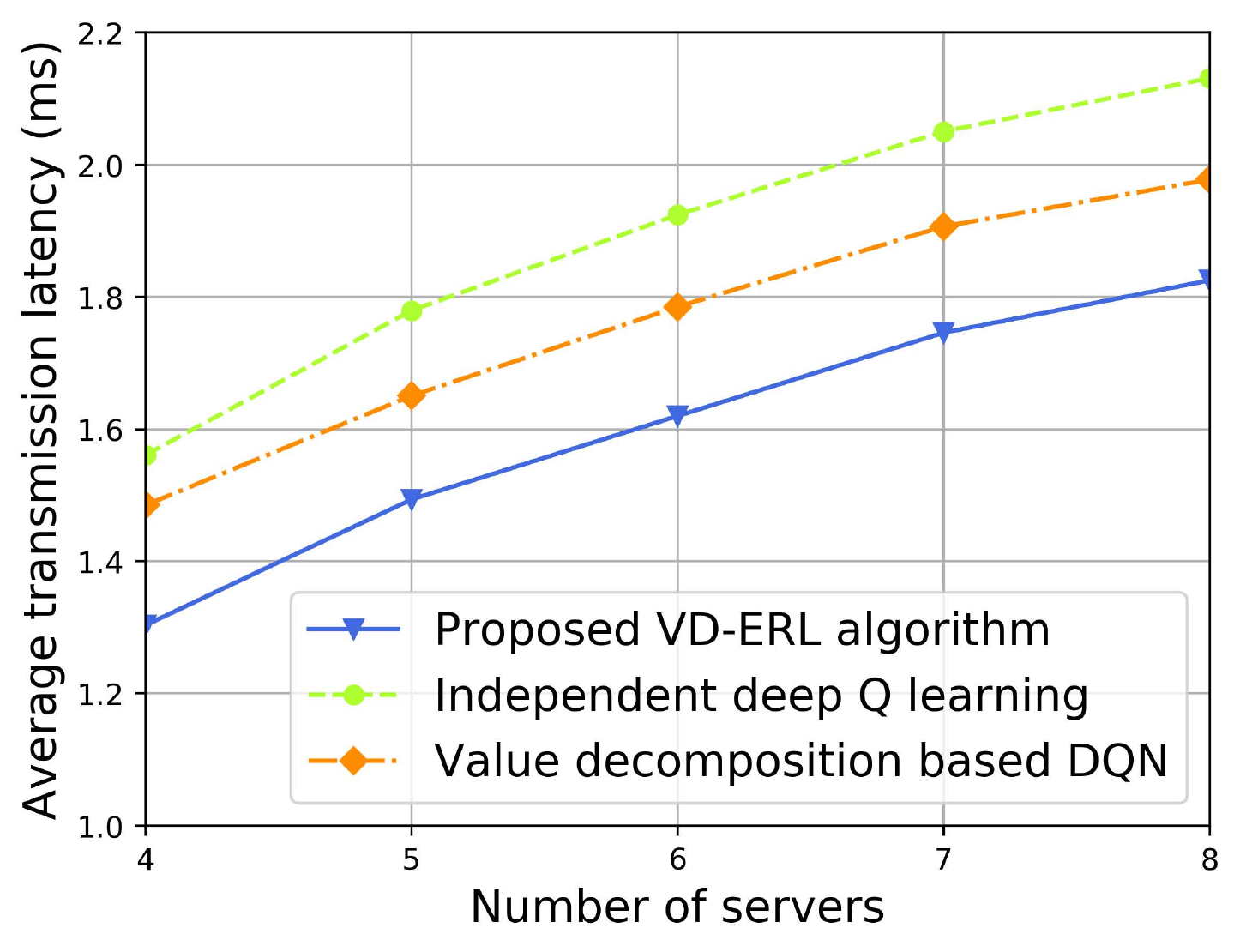}
}
\caption{Average transmission latency of the proposed VD-ERL algorithm.}
\label{fig8}
\end{figure*}
Figure 7 shows the probability that multiple servers allocate RBs to one user and this user only uses one RB from one server changes as the number of users and the number of servers varies, respectively. Hereinafter, we define the probability that multiple servers allocate RBs to one user as multi-RB allocation probability. 
From Figs. \ref{fig7}a)-\ref{fig7}d), we can see that the multi-RB allocation probability resulting from the proposed VD-ERL algorithm is 0\%, which significantly outperforms the traditional multi-agent RL algorithms. This stems from the fact that the proposed VD-ERL algorithm that aims to maximize the expected total reward enables each server to collaborate with other servers in determining RB allocation for each user thus avoiding multi-RB allocation.

Figure 8 shows the average transmission latency of all users changes as the number of users and the number of servers varies, respectively. In Figs. \ref{fig8}a) and \ref{fig8}b), we see that the average transmission latency of all considered algorithms decrease as the number of users increases. The reason is that the servers can serve the users with higher ISS using limited wireless resources. In Figs. \ref{fig8}c) and \ref{fig8}d), we can see that the average transmission latency of all considered algorithms increases as the number of servers increases. This is due to the fact that interference among users increases as the number of servers increases, and hence, the data rates of semantic information transmission decrease. From Fig.~\ref{fig8}, we can also observe that, compared to baselines a), b) and c), the proposed VD-ERL algorithm can reduce the average transmission latency by up to 74.1\%, 16.1\%, and 9.5\% respectively. This stems from the fact that the combination of entropy-maximization and VD based DRL framework enables the servers to cooperatively explore RB allocation policies to minimize transmission delay.
\begin{figure*}[t]
\centering  
\subfigure[Example 1]{
\centering
\includegraphics[width=14cm]{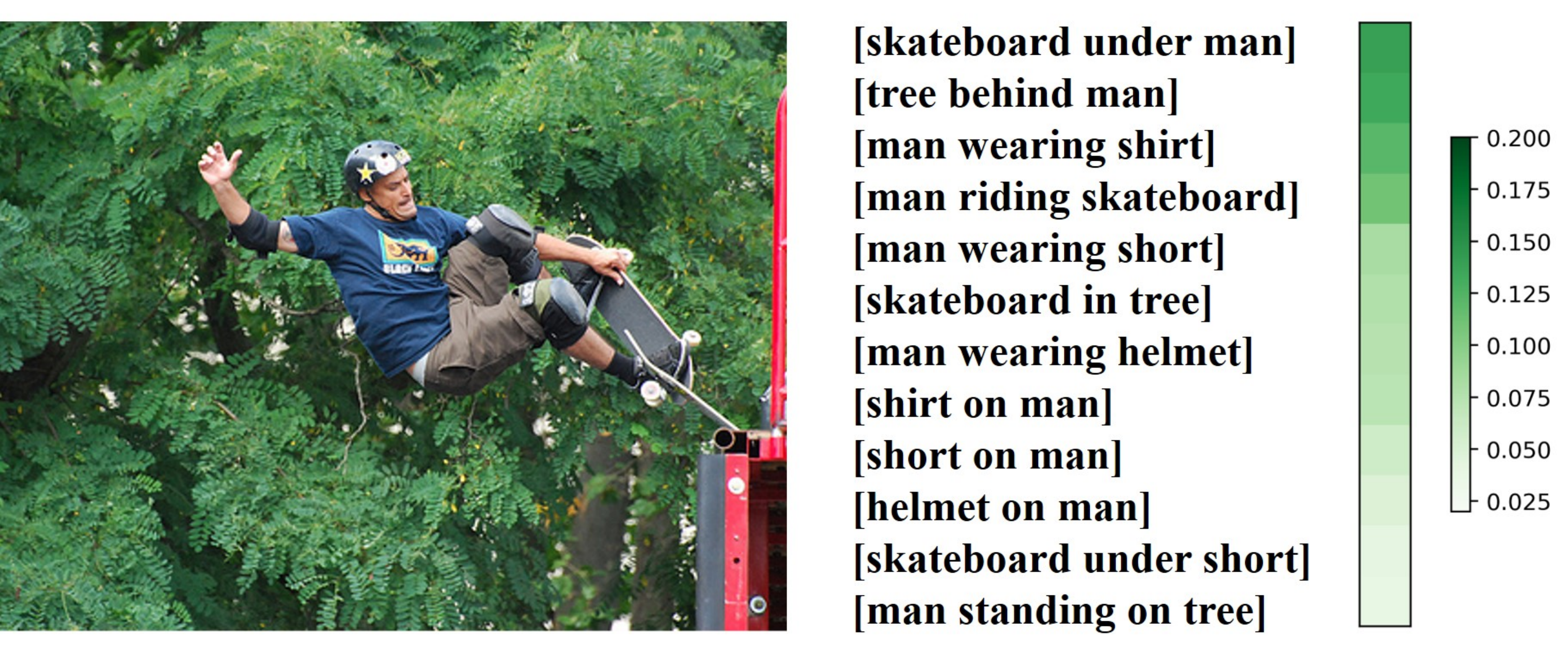}
}
\subfigure[Example 2]{
\centering
\includegraphics[width=14cm]{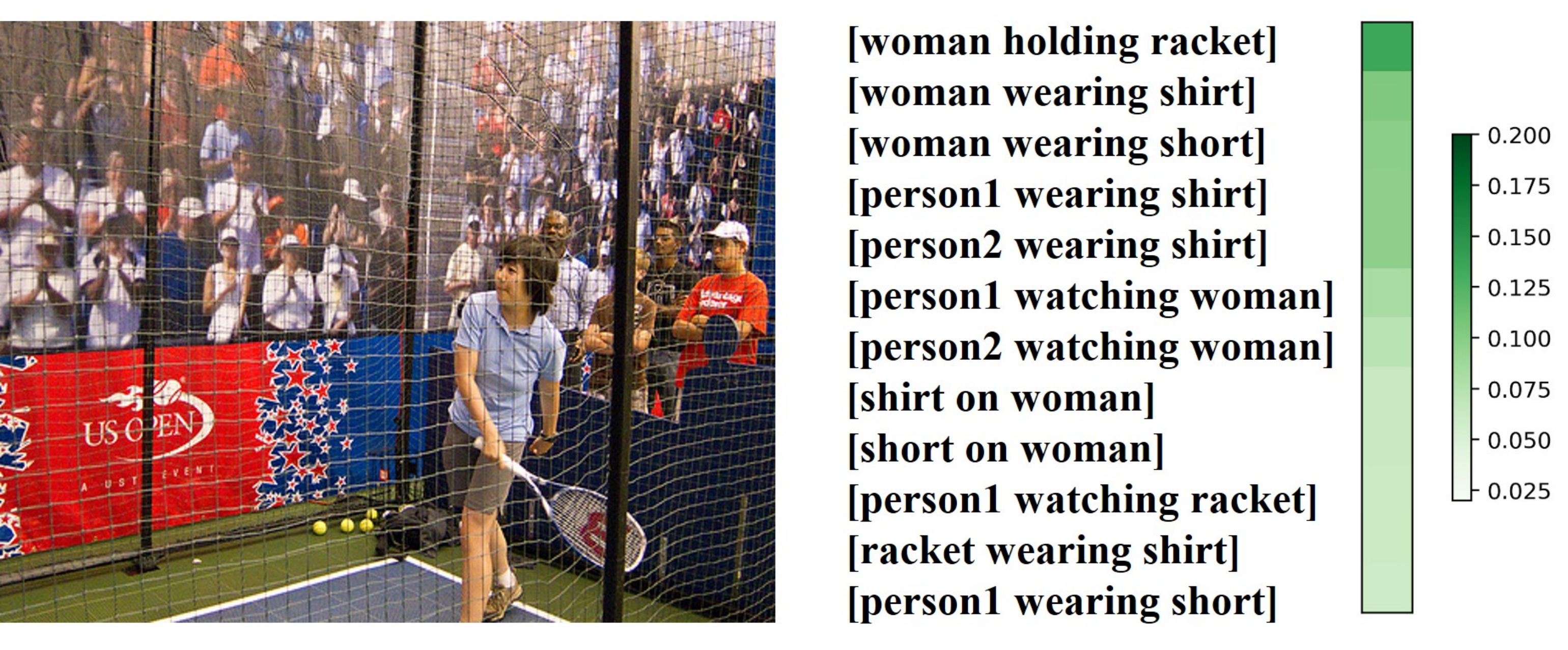}
}
\caption{The semantic scores distribution of semantic triples}
\label{fig9}
\end{figure*}

Figure 9 shows the relationships between the semantic score distribution of semantic triples and the original image. In particular, as the semantic score increases, the color of that semantic triple changes from white to green. For example, the semantic score of the greenest semantic triple ``\emph{woman holding racket}'' in Fig. 9b) is 0.1351. From Fig.~\ref{fig9}, we can see that the semantic triple with high semantic scores is more critical, e.g., the semantic triple ``\emph{man riding skateboard}'' in Fig. 9a) and the semantic triple ``\emph{woman holding racket}'' in Fig. 9b). In Fig.~\ref{fig9}, we can also see that, ranked by the semantic scores, the transmission priority of the redundant semantic triples, e.g., ``\emph{shirt on man}'' and unreasonable semantic triples, e.g., ``\emph{man standing on tree}'' is lower than other triples based on \textbf{Algorithm 1}.

\begin{figure}[tp]
\centering  
\subfigure[Semantic scores distribution of users]{
\centering
\includegraphics[width=16cm]{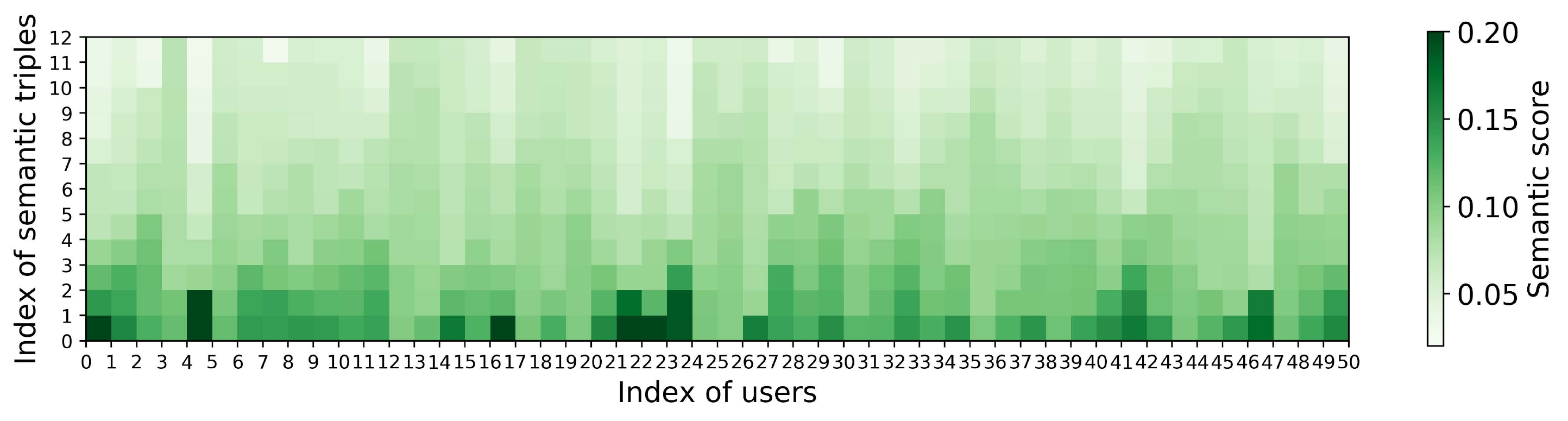}
}
\subfigure[ISS varies as the number of transmitted semantic triples varies of users]{
\centering
\includegraphics[width=16cm]{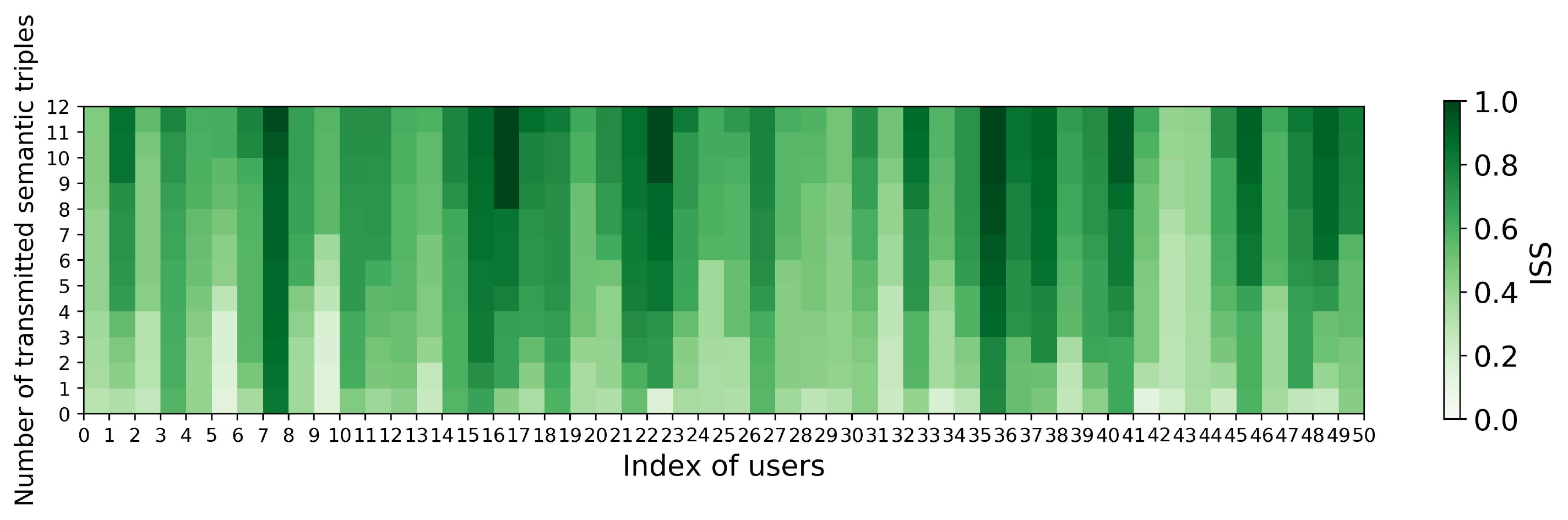}
}
\subfigure[RB allocation based on baseline b)]{
\centering
\includegraphics[width=16cm]{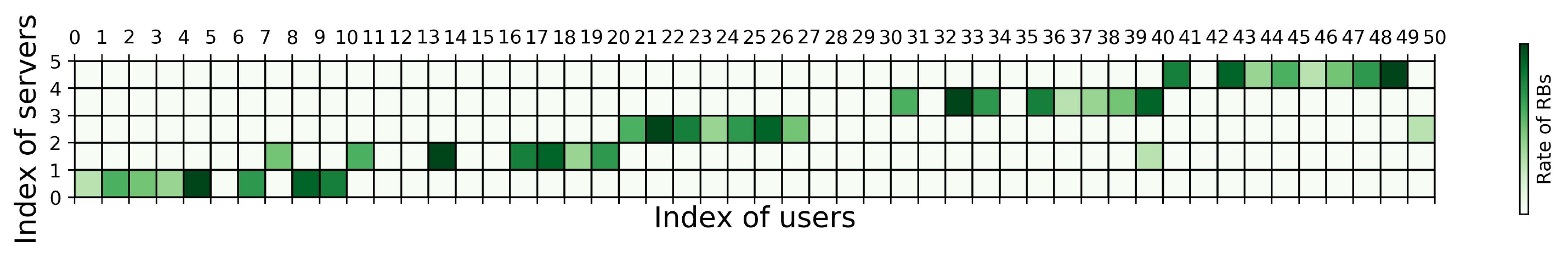}
}
\subfigure[RB allocation based on baseline c)]{
\centering
\includegraphics[width=16cm]{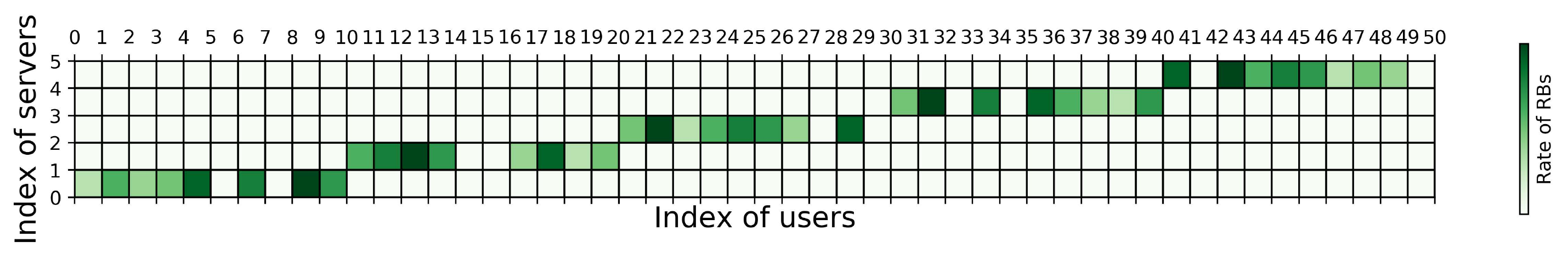}
}
\subfigure[RB allocation based on VD-ERL algorithm]{
\centering
\includegraphics[width=16cm]{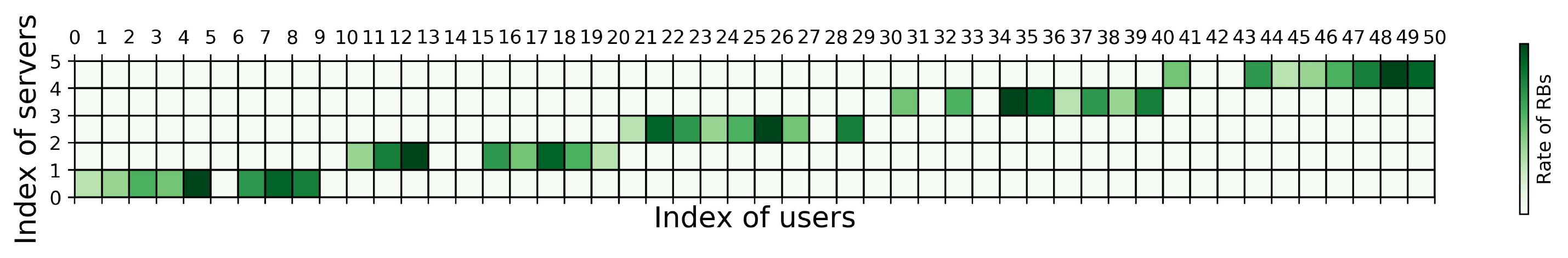}
}
\caption{Correlation between transmitted semantic information and RB allocation policy}
\label{fig10}
\end{figure}

Figure 10 shows the correlation between the transmitted semantic information and RB allocation policies of the proposed VD-ERL algorithm and baselines. In particular, in Fig. \ref{fig10}a), as the semantic score increases, the color of the semantic triple changes from white to green. Similarly, in Fig. \ref{fig10}b), the color of the transmitted partial semantic triples changes from white to green as the ISS of transmitted semantic information increases. From Fig. \ref{fig10}b), we can see that the ISS monotonically increases as the number of transmitted semantic triples increases. 
From Figs. \ref{fig10}a) and \ref{fig10}b), we can see that the semantic score can describe the importance of each semantic triple with small error.
For example, the semantic scores of the semantic information transmitted to user 5 are smaller than other semantic information and its ISS is also smaller than that of other semantic information. Figures \ref{fig10}c), \ref{fig10}d), and \ref{fig10}e) show the RB allocation results of baseline b), baseline c), and the proposed VD-ERL algorithm, respectively. In particular, the user index is determined by the distance between the user and the nearby server. A user with minimum distance will have a smallest index. For example, users 0 to 9 are close to server 1, users 10 to 19 are close to server 2, users 20 to 29 are close to server 3, and so on. In these figures, as the rate of RB increases, the color of that RB becomes greener. Then, in Figs. \ref{fig10}c), \ref{fig10}d), and \ref{fig10}e), we can see that compared to independent deep Q learning algorithm, the proposed VD-ERL algorithm enable all servers to cooperatively determine RB allocation for each user. For example, as shown in Fig. \ref{fig10}c), both server 2 and server 4 intend to allocate RBs to user 39, which causes a  multi-RB allocation problem. 

\section{Conclusion}
In this paper, we have developed a novel image semantic communication framework that enables a set of servers collaboratively transmit images to their associated users using semantic communication techniques. We have modeled the semantic information of each image as a scene graph that consists of a set of objects and relationships between them. We have proposed an ISS metric to evaluate the semantic similarity between the original image and its textual semantic information. Under the limited wireless resource constraints, each server must jointly determine the semantic information to be transmitted and the RB allocation scheme. This problem is formulated as an optimization problem whose goal is to minimize the average transmission latency while meeting the ISS requirement. To solve this problem, we have developed a value decomposition based entropy-maximized multi-agent RL algorithm that enables servers to find an optimal cooperative RBs allocation scheme based on local observation of each server. Simulation results have shown that, compared with traditional multi-agent RL algorithms, the proposed algorithm significantly reduces the transmission latency and improves the convergence speed. 
%
%
\bibliographystyle{IEEEbib}
\bibliography{ref}
\end{document}